\documentclass[review,preprint,12pt,3p]{elsarticle}

\usepackage{setspace}
\usepackage[caption=false,font=normalsize,labelfont=sf,textfont=sf]{subfig}

\usepackage{float}
\usepackage[inkscapearea=page]{svg}
\usepackage[labelsep=period]{caption}
\usepackage{balance}
\usepackage{algorithmic}
\usepackage{array}
\usepackage{url}
\usepackage{graphicx}
\usepackage{multirow}
\usepackage{verbatim}
\usepackage{pifont}
\newcommand{\cmark}{\ding{51}}%
\newcommand{\xmark}{\ding{55}}%

\usepackage{amsmath,amsfonts}

\usepackage{textcomp}
\usepackage{xcolor}
\usepackage{tabularx}
\usepackage{todonotes}
\usepackage{amssymb}
\usepackage{stfloats}

\journal{Solar Energy}
\begin{document}

\begin{frontmatter}

\title{Location-aware green energy availability forecasting for multiple time frames in smart buildings: The case of Estonia}

\author[ut1]{Mehdi Hatamian \fnref{note1}}
    \ead{mehdi.hatamian86@gmail.com}
\author[ncut]{Bivas Panigrahi}
    \ead{bivas@ncut.edu.tw}
\author[ut2]{Chinmaya~Kumar~Dehury\corref{mycorrespondingauthor}}
    \ead{chinmaya.dehury@ut.ee}
    \cortext[mycorrespondingauthor]{Corresponding author}

\address[ut1]{Independent Researcher, Tehran, Iran}

\address[ncut]{Department of Refrigeration, Air Conditioning \& Energy Engineering, National Chin-Yi University of Technology, Taichung 41170, Taiwan}
\address[ut2]{Mobile \& cloud Lab, Institute of Computer Science, University of Tartu, Tartu 50090, Estonia}

\fntext[note1]{Mr. Mehdi was with the Institute of Computer Science, University of Tartu as a Master student. Currently, he is an independent researcher.}
\date{July 2022}

\begin{abstract}

Renewable Energies (RE) have gained more attention in recent years since they offer clean and sustainable energy. One of the major sustainable development goals (SDG-7) set by the United Nations (UN) is to achieve affordable and clean energy for everyone. Among the world’s all renewable resources, solar energy is considered as the most abundant and can certainly fulfill the target of SDGs. Solar energy is converted into electrical energy through Photovoltaic (PV) panels with no greenhouse gas emissions. However, power generated by PV panels is highly dependent on solar radiation received at a particular location over a given time period. Therefore, it is challenging to forecast the amount of PV output power. Predicting the output power of PV systems is essential since several public/private institutes generate such green energy, and need to maintain the balance between demand and supply. 
This research aims to forecast PV system output power based on weather and derived features using different machine learning models. The objective is to obtain the best-fitting model to precisely predict output power by inspecting the data. Moreover, different performance metrics are used to compare and evaluate the accuracy under different machine learning models such as random forest, XGBoost, KNN, etc. 
\end{abstract}
\begin{keyword}
Solar Panel, Smart Building, Green Energy Prediction, Machine Learning, PV Output Power Prediction
\end{keyword}

\end{frontmatter}

\section{Introduction}
\label{Introduction}
To achieve sustainable development and growth, United Nations (UN) has set the blueprint for 17 sustainability development goals (SDG-17). One of the major goals (SDG-7) is to provide clean and affordable energy to the population \cite{SDG7}. Hence, the generation of Renewable Energy (RE) is strongly encouraged and supported by technological advancements and government policies for viable energy management in the future \cite{united2017sustainable, OUEDRAOGO2022120}. 
Therefore a sustainable alternative and energy management strategy would be to maximize the usage of energy produced by the Photovoltaic (PV) system \cite{safe}. Renewable energies have recently gained more attention since they offer clean and sustainable energy. Among the world's renewable resources, solar energy is the most abundant one meaning the energy is from an unlimited source that is not depleted by usage. Solar energy is converted into electrical energy through PV panels with no greenhouse gas emissions. 
Predicting the energy production by the PV system is essential since many companies generate energy, and they need to maintain electricity production and demand in balance. Moreover, an efficient way to convince investors to invest in solar energy is to provide them a time frame for a profit from their investment. However, predicting the output power generated by PV systems is a cumbersome task since they are highly dependent on how much solar radiation they receive, the condition of weather, the position of the PV panel, and the amount of time PV panels are exposed to sunlight, and many more \cite{ZHU2022119269}. Solar radiation is crucial for PV systems, and the output power of the PV system is determined by total solar irradiance on a particular day. A unit of irradiance is defined by the total output of the solar source falling on a unit area. However, solar irradiance is affected by various  factors, including weather, location, time, etc.

Therefore, a direct relationship between energy produced by a PV system and local weather conditions exists that varies during the day as the amount of solar irradiance changes \cite{rodriguez}. Furthermore, predicting the amount of electricity generated by a PV system is crucial to calculating the size of the system, system load measurements, and return on investment (ROI) \cite{ROI,JEBLI2021120109}. 
Such varied parameters play a major role in making the energy production prediction more complex \cite{sharma}.


Different methods have been employed in the literature to predict the output power of PV systems. Data-driven and model-based are commonly used methods for predicting green energy generation \cite{model}. While model-based methods rely on analytical equations by leveraging meteorological weather data \cite{model_based}, the data-driven models utilize machine learning techniques to predict the output power of the PV system. However, to meet the demand for modern PV systems and a sustainable energy management strategy, the existing prediction approaches are not sufficient enough. 

This paper aims to predict the output power of PV systems using state-of-art machine learning models, including Extreme Gradient Boosting (XGboost), Random Forest (RF), K-Nearest Neighbors (KNN), Support Vector Regression (SVR), and Multi-Layer Perceptron (MLP). We have investigated the effect of meteorological data on predicted output power to find an optimal set of input features. Moreover, the overall impact of three derived features, including ``Hours," ``Month," and ``Prior Output Power," is analyzed. The list of acronyms is presented in Table \ref{tab:acronym}.

\subsection{Motivations and Goals}
High penetration of PV systems is offered as an alternative to energy production methods due to its economic benefits and sustainable clean energy. However, the stability of the PV systems might be threatening without an accurate prediction of the PV energy production. The energy production of the PV systems is dependent on meteorological data. Therefore to maintain the stability of the PV system, the uncertainty of output power predictions must be addressed by accurate forecasting or prediction tools. The state-of-the-art prediction model relies mainly on the historical performance of the PV panels without taking advantage of the cloud coverage and other weather-related information. To meet the rising demand for a futuristic green energy eco-system, a prediction model needs to consider the location and weather. It is observed from the state-of-the-art literature survey that a prediction model may not work in all geographical regions. Further, a single model may not give the desired accuracy throughout the year for all the sessions, which is one of the major motivational situations for this work.

The selection of ML models is another motivational scenario behind this proposal. Instead of relying on a single ML model, it is necessary to investigate and compare the performance and accuracy of different ML models, such as SVR, RF, and XGBoost, in different situations. The performance needs to be monitored based on various metrics, such as Mean Absolute Percentage Error (MAPE), Mean Absolute Error (MAE), Root Mean Square Error (RMSE), and R-squared. These aforementioned research challenges motivate us to present a location-based green energy availability in a smart building located in Tartu city of Estonia.



\subsection{Contributions}

The main contributions of this research study to the field of predicting PV output power can be summarised as follows:

\begin{itemize}
    \item The research challenge of predicting the PV energy output is investigated with the real datasets and recent research results. 
    \item A real historical dataset of 1 yr duration from the solar panels installed on the roof-top of the university building and nearby weather station in Estonia are collected and pre-processed.
    \item Transformation is introduced as an efficient way to normalize the dependent variables to alter the skewness of the data and remove or lessen the impact of seasonality and trend in our data.  
    \item Z-score, Pearson correlation, and permutation-based feature importance are proposed to be applied to the input features as a feature scaling method.
    \item Five popular ML models: KNN, XGBoost, MLP, SVR, and RF, are implemented, and the performance results are compared based on MAPE, MAE, R2, and RMSE. 
    \item The importance of the previous output power on the prediction model is analyzed. 
\end{itemize}

The rest of this research paper is structured as follows. In Section \ref{literature}, an overview of general research methods is presented, followed by the methodology described in Section \ref{methodology}. Section \ref{dataprocessing} contains data processing, including target transformation, outlier handling, and feature selection. In Section \ref{implementation}, the details about the implementation of each model are given. The importance of prior output power is given in Section \ref{prior}. Moreover, in Section \ref{results}, results are compared and discussed in detail for each ML model, followed by the concluding remarks and future works in Section \ref{conclusion}.

\begin{table}[!h]
    \centering
    \caption{List of Acronyms} \label{tab:acronym}
    \footnotesize
    \def\arraystretch{1.1}
    \begin{tabular}{|c|c|}
        \hline
        \textbf{Acronyms} & \textbf{Description} \\ \hline
        ANN  &  Artificial Neural Network\\ \hline
        AR  &  Adaptive Recursive Linear\\ \hline
        CV  &  Cross Validation\\ \hline
        FFBP  &  Feed-forward Back Propagation\\ \hline
        GBRT  &  Gradient Boosted Regression Trees\\ \hline
        GHI  &  Global Horizontal Irradiance\\ \hline
        GRNN  &  General Regression Neural Network\\ \hline
        IQR  &  Interquartile Range\\ \hline
        KNN  &  K-Nearest Neighbour\\ \hline
        MAE  &  Mean Absolute Error\\ \hline
        MAPE  &  Mean Absolute Percentage Error\\ \hline
        ML  &  Machine Learning\\ \hline
        MLP  &  Multilayer Perceptron\\ \hline
        NWP  &  Numerical Weather Prediction\\ \hline
        PV  &  Photovoltaic\\ \hline
        RBF  &  Radial Basis Function\\ \hline
        RE  &  Renewable Energy\\ \hline
        RF  &  Random Forest\\ \hline
        RMSE  &  Root Mean Square Error\\ \hline
        ROI  &  Return on Investment\\ \hline
        RT  &  Regression Trees\\ \hline
        SGD  &  Stochastic Gradient Descent\\ \hline
        STD  &  Standard Deviation\\ \hline
        SVR  &  Support Vector Regression\\ \hline
        XGBoost  &  Extreme Gradient Boosting\\ \hline
    \end{tabular}
\end{table}

\section{Literature Review}\label{literature}

This section provides insight into general research and information about the state-of-the-art literature on PV output power prediction.

\subsection{Review research on solar energy power prediction}

Khatib et al. \cite{khatib} presented a literature review of solar energy modeling techniques. An overview of solar energy modeling techniques is presented that are categorized by their nature. The author concluded that the sunshine ratio, ambient temperature, and relative humidity show the strongest correlation with solar energy. However, the author reported that the accuracy of the predicted value is the most important challenge regarding solar energy modeling. Another problematic challenge is measuring historical weather data since weather stations have different comprehensive measuring devices. Similarly, Inman et al. \cite{inman} conducted a literature study to provide a broad understanding of theories and predicting techniques regarding PV systems. Theories from different fields are discussed, including solar irradiance, air masses, and clearness indices, and the study serves as guidance for physic, statistical, and ML perspective. However, a number of major challenges are identified, including unpredictable and steep ramps, making up errors in forecasting output power, intra-hour variability, and over generation in the middle of the night.

\subsection{Machine learning Models}
Chuluunsaikhan et al. \cite{japa} presented separate analyses of the effect of weather and air pollution on PV output power. This research paper compared several ML models, including SVR, MLP, KNN, RF, and Gradient Boosting. Additionally, various sources are suggested for the predictions, such as meteorological, solar panel, and air pollution features. The author concluded that RF has better accuracy than the other models. However, the suggested method for feature selection is based on Pearson correlation, while only using Pearson Correlation is not reliable for the field of the PV system. Additionally, some of the details on time frame and implementation are missing.

Another research was conducted by J.Barrera et al. \cite{barrera} using Artificial Neural Network (ANN) to study how various factors impact the prediction of the output power. By evaluating models based on lower MSE, the author claims have gained more accuracy compared to other literature proposals. However, while relative humidity is one of the most important factors that has a stronger impact on overall output power, the contribution of the relative humidity is neglected by the author. Moreover, the predictions are performed in 5 minutes intervals that the variation in weather is not noticeable in very short time frames. 
Additionally, Theocharides et al. \cite{theocharides} proposed a comparison-based study using several ML models, including ANNs, SVR, and Regression Trees (RTs). The authors aim to examine different hyperparameters and a set of features to predict PV system output power. Finally, the author claims that ANN outperforms other models by predicting the accurate output power of the PV system. However, the underlying method used for data processing is not clear. Moreover, to boost the accuracy of prediction, khademi et al. \cite{khademi} suggested an ANN model and splitting the weather into sunny and cloudy days. However, this suggestion is highly dependent on the geographical location since the pattern of weather in different seasons may be similar regardless of the seasonal changes. 

A comparison between SVR and an analytical model is proposed by Nageem et al. \cite{nageem}. In the analytical method, the produced output power of the PV system is predicted by solar panel orientation, location, and solar irradiance. 
The author uses hourly measured parameters as input features and concludes that the accuracy of analytical models is slightly less than SVR. Moreover, in research conducted by De Leone et al. \cite{leone} the author uses the past meteorological data to predict the future output power of PV systems by SVR. The author concludes that the quality of meteorological weather data is crucial and determines the accuracy of the predicted output power. However, the selected time interval is unreliable since the meteorological data variation is not noticeable during 15 minutes of intervals. 

Additionally, Shi et al. \cite{shi} propose a study to use SVR to forecast the output power of a PV system. The study aims to propose a classification of the meteorological data for correlation analysis on the local weather and the output power predictions. Radial Basis Function (RBF) is used as the kernel to fit the different weather classes by finding a way to train data on a given weather condition. Persson et al. \cite{persson} conducted a study to use Gradient Boosted Regression Trees (GBRT) for forecasting output power by leveraging historical meteorological data and output power data. The study aims to predict solar energy production for 1-6 hours ahead. The author reports that GBRT outperforms time-series, adaptive recursive linear (AR), and climatology models. However, GBRT has no simple updating procedure, but when sufficient new data is available, the model can be re-fitted, thus yielding more efficiency.

\subsection{Literature summary and Insights}




\newcommand{\STAB}[1]{\begin{tabular}{@{}c@{}}#1\end{tabular}}

\begin{table*}
\centering
\caption{Comparison of state-of-the-arts based on features, ML models and their accuracy}
\label{tbl:featuresw}
\scriptsize
\def\arraystretch{1.2}
\begin{tabular}{p{0.15\textwidth}c|cccccc|}
\cline{3-8}
                                                 &                                                                                   & \multicolumn{6}{c|}{\textbf{Research Papers}}                                                                                                                                                                                                                                                                                                                                                                                                                                                                                                    \\ \cline{3-8} 
                                                 &                                                                                   & \multicolumn{1}{c|}{\textbf{\cite{japa}}} & \multicolumn{1}{c|}{\textbf{\cite{model}}} & \multicolumn{1}{c|}{\textbf{\cite{barrera}}} & \multicolumn{1}{c|}{\textbf{\cite{nageem}}} & \multicolumn{1}{c|}{\textbf{\cite{theocharides}}} & \textbf{ \cite{leone}} \\ \hline
\multicolumn{1}{|c|}{\multirow{18}{*}{\STAB{\rotatebox[origin=c]{90}{Popular Features}}}} & \textbf{\begin{tabular}[c]{@{}c@{}}Tilt Angle \\ and Orientation\end{tabular}}    & \multicolumn{1}{c|}{\cmark}                                                     & \multicolumn{1}{c|}{\xmark}                                                & \multicolumn{1}{c|}{\cmark}                                             & \multicolumn{1}{c|}{\xmark}                                            & \multicolumn{1}{c|}{\xmark}                                                   & \xmark                                               \\ \cline{2-8} 
\multicolumn{1}{|c|}{}                           & \textbf{\begin{tabular}[c]{@{}c@{}}Module \\ Temperature\end{tabular}}            & \multicolumn{1}{c|}{\cmark}                                                     & \multicolumn{1}{c|}{\xmark}                                                & \multicolumn{1}{c|}{\xmark}                                             & \multicolumn{1}{c|}{\xmark}                                            & \multicolumn{1}{c|}{\xmark}                                                   & \xmark                                               \\ \cline{2-8} 
\multicolumn{1}{|c|}{}                           & \textbf{\begin{tabular}[c]{@{}c@{}}Direct Normal \\ Irradiance\end{tabular}}      & \multicolumn{1}{c|}{\xmark}                                                     & \multicolumn{1}{c|}{\xmark}                                                & \multicolumn{1}{c|}{\cmark}                                             & \multicolumn{1}{c|}{\xmark}                                            & \multicolumn{1}{c|}{\xmark}                                                   & \xmark                                               \\ \cline{2-8} 
\multicolumn{1}{|c|}{}                           & \textbf{\begin{tabular}[c]{@{}c@{}}Diffuse Horizontal \\ Irradiance\end{tabular}} & \multicolumn{1}{c|}{\xmark}                                                     & \multicolumn{1}{c|}{\xmark}                                                & \multicolumn{1}{c|}{\cmark}                                             & \multicolumn{1}{c|}{\cmark}                                            & \multicolumn{1}{c|}{\xmark}                                                   & \xmark                                               \\ \cline{2-8} 
\multicolumn{1}{|c|}{}                           & \textbf{\begin{tabular}[c]{@{}c@{}}Global Horizontal \\ Irradiance\end{tabular}}  & \multicolumn{1}{c|}{\cmark}                                                     & \multicolumn{1}{c|}{\cmark}                                                & \multicolumn{1}{c|}{\xmark}                                             & \multicolumn{1}{c|}{\xmark}                                            & \multicolumn{1}{c|}{\cmark}                                                   & \cmark                                               \\ \cline{2-8} 
\multicolumn{1}{|c|}{}                           & \textbf{Reflected Irradiation}                                                    & \multicolumn{1}{c|}{\xmark}                                                     & \multicolumn{1}{c|}{\xmark}                                                & \multicolumn{1}{c|}{\cmark}                                             & \multicolumn{1}{c|}{\xmark}                                            & \multicolumn{1}{c|}{\xmark}                                                   & \xmark                                               \\ \cline{2-8} 
\multicolumn{1}{|c|}{}                           & \textbf{Humidity}                                                                 & \multicolumn{1}{c|}{\cmark}                                                     & \multicolumn{1}{c|}{\cmark}                                                & \multicolumn{1}{c|}{\xmark}                                             & \multicolumn{1}{c|}{\cmark}                                            & \multicolumn{1}{c|}{\cmark}                                                   & \xmark                                               \\ \cline{2-8} 
\multicolumn{1}{|c|}{}                           & \textbf{Sunshine}                                                                 & \multicolumn{1}{c|}{\cmark}                                                     & \multicolumn{1}{c|}{\xmark}                                                & \multicolumn{1}{c|}{\xmark}                                             & \multicolumn{1}{c|}{\xmark}                                            & \multicolumn{1}{c|}{\xmark}                                                   & \xmark                                               \\ \cline{2-8} 
\multicolumn{1}{|c|}{}                           & \textbf{Cloud Coverage}                                                           & \multicolumn{1}{c|}{\cmark}                                                     & \multicolumn{1}{c|}{\xmark}                                                & \multicolumn{1}{c|}{\xmark}                                             & \multicolumn{1}{c|}{\xmark}                                            & \multicolumn{1}{c|}{\xmark}                                                   & \xmark                                               \\ \cline{2-8} 
\multicolumn{1}{|c|}{}                           & \textbf{Temperature}                                                              & \multicolumn{1}{c|}{\cmark}                                                     & \multicolumn{1}{c|}{\cmark}                                                & \multicolumn{1}{c|}{\cmark}                                             & \multicolumn{1}{c|}{\cmark}                                            & \multicolumn{1}{c|}{\cmark}                                                   & \cmark                                               \\ \cline{2-8} 
\multicolumn{1}{|c|}{}                           & \textbf{Wind Speed}                                                               & \multicolumn{1}{c|}{\xmark}                                                     & \multicolumn{1}{c|}{\cmark}                                                & \multicolumn{1}{c|}{\cmark}                                             & \multicolumn{1}{c|}{\cmark}                                            & \multicolumn{1}{c|}{\cmark}                                                   & \xmark                                               \\ \cline{2-8} 
\multicolumn{1}{|c|}{}                           & \textbf{Snowfall}                                                                 & \multicolumn{1}{c|}{\xmark}                                                     & \multicolumn{1}{c|}{\xmark}                                                & \multicolumn{1}{c|}{\xmark}                                             & \multicolumn{1}{c|}{\xmark}                                            & \multicolumn{1}{c|}{\xmark}                                                   & \xmark                                               \\ \cline{2-8} 
\multicolumn{1}{|c|}{}                           & \textbf{Precipitation}                                                            & \multicolumn{1}{c|}{\xmark}                                                     & \multicolumn{1}{c|}{\xmark}                                                & \multicolumn{1}{c|}{\xmark}                                             & \multicolumn{1}{c|}{\xmark}                                            & \multicolumn{1}{c|}{\xmark}                                                   & \xmark                                               \\ \cline{2-8} 
\multicolumn{1}{|c|}{}                           & \textbf{Atmospheric Pressure}                                                     & \multicolumn{1}{c|}{\xmark}                                                     & \multicolumn{1}{c|}{\xmark}                                                & \multicolumn{1}{c|}{\xmark}                                             & \multicolumn{1}{c|}{\cmark}                                            & \multicolumn{1}{c|}{\xmark}                                                   & \xmark                                               \\ \cline{2-8} 
\multicolumn{1}{|c|}{}                           & \textbf{Sunlight}                                                                 & \multicolumn{1}{c|}{\xmark}                                                     & \multicolumn{1}{c|}{\xmark}                                                & \multicolumn{1}{c|}{\xmark}                                             & \multicolumn{1}{c|}{\xmark}                                            & \multicolumn{1}{c|}{\xmark}                                                   & \xmark                                               \\ \cline{2-8} 
\multicolumn{1}{|c|}{}                           & \textbf{Daylight}                                                                 & \multicolumn{1}{c|}{\xmark}                                                     & \multicolumn{1}{c|}{\xmark}                                                & \multicolumn{1}{c|}{\xmark}                                             & \multicolumn{1}{c|}{\xmark}                                            & \multicolumn{1}{c|}{\xmark}                                                   & \xmark                                               \\ \cline{2-8} 
\multicolumn{1}{|c|}{}                           & \textbf{Wind Direction}                                                           & \multicolumn{1}{c|}{\xmark}                                                     & \multicolumn{1}{c|}{\xmark}                                                & \multicolumn{1}{c|}{\xmark}                                             & \multicolumn{1}{c|}{\xmark}                                            & \multicolumn{1}{c|}{\cmark}                                                   & \xmark                                               \\ \cline{2-8} 
\multicolumn{1}{|c|}{}                           & \textbf{Azimuth Angle}                                                            & \multicolumn{1}{c|}{\xmark}                                                     & \multicolumn{1}{c|}{\xmark}                                                & \multicolumn{1}{c|}{\xmark}                                             & \multicolumn{1}{c|}{\xmark}                                            & \multicolumn{1}{c|}{\cmark}                                                   & \xmark                                               \\ \hline
\\ \hline
\multicolumn{1}{|c|}{\multirow{7}{*}{\STAB{\rotatebox{90}{\parbox{5em}{\textbf{Most widely used ML model}}}}}} & \textbf{SVR}               & \multicolumn{1}{c|}{\cmark}                                                     & \multicolumn{1}{c|}{\xmark}                                                & \multicolumn{1}{c|}{\xmark}                                             & \multicolumn{1}{c|}{\cmark}                                            & \multicolumn{1}{c|}{\cmark}                                                   & \cmark                                               \\ \cline{2-8} 
\multicolumn{1}{|c|}{}                                         & \textbf{KNN}               & \multicolumn{1}{c|}{\cmark}                                                     & \multicolumn{1}{c|}{\xmark}                                                & \multicolumn{1}{c|}{\xmark}                                             & \multicolumn{1}{c|}{\xmark}                                            & \multicolumn{1}{c|}{\xmark}                                                   & \xmark                                               \\ \cline{2-8} 
\multicolumn{1}{|c|}{}                                         & \textbf{RF}                & \multicolumn{1}{c|}{\cmark}                                                     & \multicolumn{1}{c|}{\xmark}                                                & \multicolumn{1}{c|}{\xmark}                                             & \multicolumn{1}{c|}{\xmark}                                            & \multicolumn{1}{c|}{\xmark}                                                   & \xmark                                               \\ \cline{2-8} 
\multicolumn{1}{|c|}{}                                         & \textbf{GB}                & \multicolumn{1}{c|}{\cmark}                                                     & \multicolumn{1}{c|}{\xmark}                                                & \multicolumn{1}{c|}{\xmark}                                             & \multicolumn{1}{c|}{\xmark}                                            & \multicolumn{1}{c|}{\xmark}                                                   & \xmark                                               \\ \cline{2-8} 
\multicolumn{1}{|c|}{}                                         & \textbf{Linear Regression} & \multicolumn{1}{c|}{\cmark}                                                     & \multicolumn{1}{c|}{\xmark}                                                & \multicolumn{1}{c|}{\xmark}                                             & \multicolumn{1}{c|}{\xmark}                                            & \multicolumn{1}{c|}{\xmark}                                                   & \xmark                                               \\ \cline{2-8} 
\multicolumn{1}{|c|}{}                                         & \textbf{ANN}               & \multicolumn{1}{c|}{\cmark}                                                     & \multicolumn{1}{c|}{\cmark}                                                & \multicolumn{1}{c|}{\cmark}                                             & \multicolumn{1}{c|}{\xmark}                                            & \multicolumn{1}{c|}{\cmark}                                                   & \xmark                                               \\ \cline{2-8} 
\multicolumn{1}{|c|}{}                                         & \textbf{Regression Tree}   & \multicolumn{1}{c|}{\xmark}                                                     & \multicolumn{1}{c|}{\cmark}                                                & \multicolumn{1}{c|}{\cmark}                                             & \multicolumn{1}{c|}{\xmark}                                            & \multicolumn{1}{c|}{\cmark}                                                   & \xmark                                               \\ \hline
\\ \hline
\multicolumn{1}{|c|}{\multirow{5}{*}{\STAB{\rotatebox{90}{\parbox{5em}{\textbf{Accuracy of most popular ML models}}}}}} & \textbf{RMSE} & \multicolumn{1}{c|}{2.38}                                                                               & \multicolumn{1}{c|}{-}                                                                       & \multicolumn{1}{c|}{0.2}                                                                         & \multicolumn{1}{c|}{-}                                                                          & \multicolumn{1}{c|}{-}                                                                                & 0.99                                                                         \\ \cline{2-8} 
\multicolumn{1}{|c|}{}                                              & \textbf{MSE}  & \multicolumn{1}{c|}{-}                                                                                  & \multicolumn{1}{c|}{-}                                                                            & \multicolumn{1}{c|}{0.04}                                                                        & \multicolumn{1}{c|}{-}                                                                          & \multicolumn{1}{c|}{-}                                                                                & -                                                                            \\ \cline{2-8} 
\multicolumn{1}{|c|}{}                                              & \textbf{MAE}  & \multicolumn{1}{c|}{1.38}                                                                               & \multicolumn{1}{c|}{6.75}                                                                        & \multicolumn{1}{c|}{0.161}                                                                       & \multicolumn{1}{c|}{-}                                                                          & \multicolumn{1}{c|}{-}                                                                                & -                                                                            \\ \cline{2-8} 
\multicolumn{1}{|c|}{}                                              & \textbf{R2}   & \multicolumn{1}{c|}{0.87}                                                                               & \multicolumn{1}{c|}{-}                                                                            & \multicolumn{1}{c|}{-}                                                                           & \multicolumn{1}{c|}{-}                                                                          & \multicolumn{1}{c|}{-}                                                                                & 0.95                                                                         \\ \cline{2-8} 
\multicolumn{1}{|c|}{}                                              & \textbf{MAPE} & \multicolumn{1}{c|}{-}                                                                                  & \multicolumn{1}{c|}{-}                                                                            & \multicolumn{1}{c|}{-}                                                                           & \multicolumn{1}{c|}{0.36}                                                                       & \multicolumn{1}{c|}{0.6}                                                                              & 0.35                                                                         \\ \hline 
\end{tabular}
\end{table*}

As a result of the above research works, it is evident that efforts have primarily been dedicated to improving the data-driven prediction models and examining other advanced models from the literature. In most research papers, the underlying data processing and implementation methods are not discussed well enough. All studies have demonstrated the importance of meteorological data in which the models are very sensitive to the recorded weather data. To address this challenge, in \cite{isak} the author suggests using different sources for historical weather data. Most of the research papers discuss the importance of non-linear models because of their ability to generalize better. 

Finally, the most relevant studies are compared based on the used ML models, features, and obtained accuracy in terms of several performance metrics, as shown in Table \ref{tbl:featuresw}. The most used features are temperature, wind speed, Global Horizontal Irradiance (GHI), and humidity, as shown in the first part of Table \ref{tbl:featuresw}. However, to the best of our knowledge and based on the literature survey, some meteorological features, including snowfall, precipitation, daylight, and sunlight, are not used at all. 

The second part of Table \ref{tbl:featuresw} summarizes the ML models used in different research works. It is observed that SVR and ANN are the most widely used ML models, followed by regression trees. From the survey, it is also observed that KNN, RF, GB, and Linear regression are some of the candidate ML models that need to be considered while improving the prediction accuracy, as given in Table \ref{tbl:featuresw}. For the further in-depth survey, the last part of Table \ref{tbl:featuresw} presents the results (in terms of RMSE, MSE, MAE, R2, and MAPE) from the selected research papers. Although in \cite{japa}, the best accuracy has been obtained by RF, but ANN has shown the best accuracy in the study conducted by \cite{model}, \cite{barrera}, and \cite{theocharides}. Moreover, compared to analytical models, SVR tends to outperform in the study conducted by \cite{leone}.

\section{Methodology} \label{methodology}

\begin{figure*}[h]
    \centering
    \includegraphics{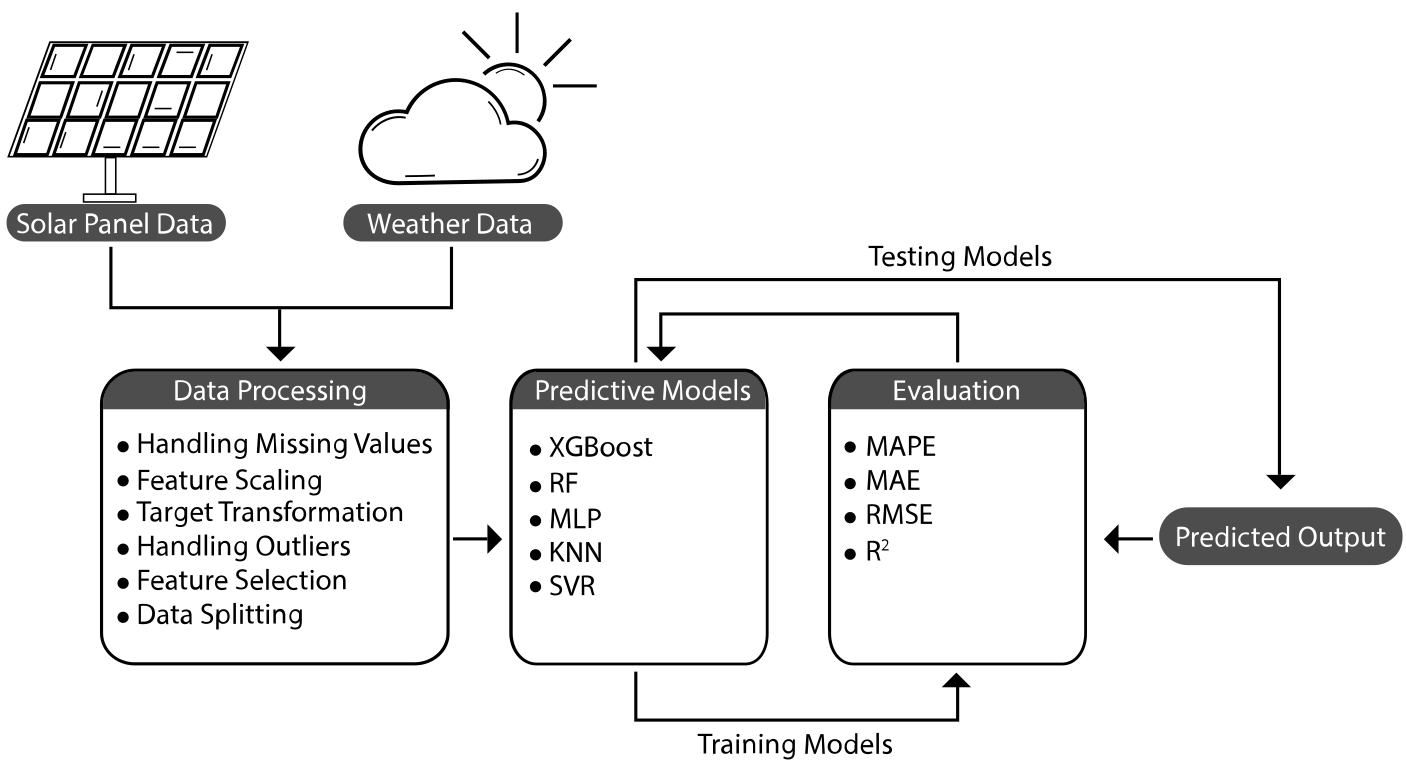}
    \caption{Model Architecture for the entire data acquisition to prediction journey.}
    \label{fig:architecture}
\end{figure*}

A description of the research methodology, including ML models and data acquisition, is presented in this section. The overall model architecture is given in Figure \ref{fig:architecture}, which represents the entire end-to-end journey of the data from acquisition to prediction. As shown in Figure \ref{fig:architecture}, the historical data are collected from PV panels and the nearby weather station. This is followed by data processing, including handling missing values, feature scaling and selection, target transformation, data splitting, etc. Several ML models, including XGBoost, RF, MLP, KNN, and SVR, are applied to a portion of the processed data to train the model, and based on the observed performance metrics (especially MAPE, MAE, RMSE, and R2), several iterations are performed. The model is then tested on the rest of the data for further improvement.

\subsection{Machine Learning Models}
ML models are used to predict the output power of PV systems by utilizing historical weather and output power data.
RF and XGBoost are selected as the tree-based algorithms. RF seems to have a high accuracy rate, based on previous studies. 
In addition, XGBoost was selected, which is rarely used in the field of output power forecasting but gained popularity due to its high performance among tree-based algorithms.  

ANN performs well and has shown great results in the field of predicting. Thus, MLP was selected as a class of ANN to predict the output power of the PV system. Because the historical meteorological data is highly correlated to their neighbors, KNN was taken into consideration in this study. More specifically, the neighbors closer to each other contribute more than those farther away. SVR has shown great accuracy in nonlinearity modeling. Therefore, SVR was selected because of the non-linearity in meteorological data. 

\subsection{Data Acquisition}
\begin{figure}[h] 
\centering
\captionsetup{width=.8\linewidth}
\includegraphics[width=0.8\linewidth]{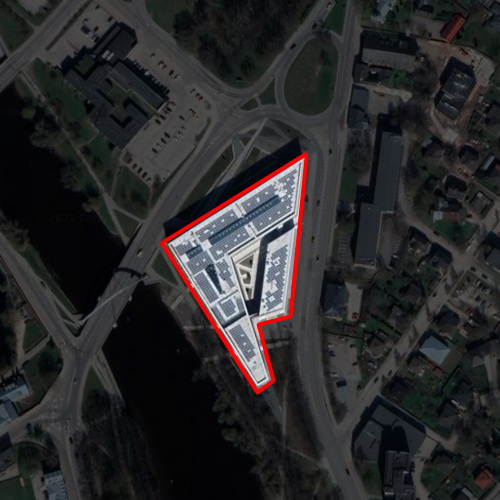}
\caption{Delta Center PV Map (retrieved and adapted from Google Maps. src: https://goo.gl/maps/1hYAXXVQSaqT7p4c6}

\label{fig:tartu}
\end{figure}

The PV output power data was collected from the University of Tartu\footnote{\url{https://ut.ee/en/}} between 20-9-2020 and 17-10-2021, at 5-minutes intervals, in \emph{XLS} format. The data contains 46268 data points (or records). The Table \ref{tbl:solar_data} represents the information about output power data.
This research paper presents a real case study of a PV system installed on the rooftop of the Delta building\footnote{\url{https://delta.ut.ee/en/}} owned by the University of Tartu in Estonia, illustrated in Figure \ref{fig:tartu}.

\begin{table}
    \centering
    \caption{Schema of the solar data }
    \label{tbl:solar_data}
    \footnotesize  
    \setlength{\tabcolsep}{2pt} 
    \begin{tabular}{|p{0.13\linewidth}|p{0.4\linewidth}|p{0.2\linewidth}|p{0.2\linewidth}|}
    \hline
    \textbf{Header}   & \textbf{Description}    & \textbf{Unit}      & \textbf{Range} \\ \hline
     ID               & ID of the solar panel                                       & NA                 & NA             \\ \hline
    DateTime              & Timestamp of the observation                            & mm/dd/yyyy  hh:mm:ss &$09/20/2020 7:38:29 PM$ PM to $10/17/2021 10:05:16 PM$             \\ \hline
    Value                 & Measured accumulative output power of PV panels         & kWh                &$-3600$ to $+365365$           \\ \hline
    Unit                  & Determines how output power is measured                 & NA                & NA             \\ \hline
    \end{tabular}
\end{table}

The meteorological data obtained from Tartu-Tõravere weather station\footnote{\url{https://euweather.eu/stations.php?lat=58.2639&lon=26.4614&lang=en}} located at a distance of approximately 5km from the Delta building. The Meteorological data is provided by the Republic of Estonia Environment Agency\footnote{\url{http://www.ilmateenistus.ee}} between 20-9-2020 and 27-9-2021 in \emph{XLS} format with 15312 data points (or records). 
The meteorological data consists of ten variables collected hourly given, as presented in Table \ref{tab:weather_variables}. Several important features are evaluated for the final predictions, including, Relative Humidity, Temperature, and GHI. The table also provides the description of the features, their units, and the range of the values. The range here refers to the corresponding minimum and maximum values present in the dataset.
The meteorological data is collected at an interval of one hour (unlike solar panel data, which is collected at an interval of 5-minutes). The collected PV panel's data and weather-related data can be accessed from the public Github Repository at \cite{data_githubrepo}.

\begin{table*}
\centering
\caption{List of features related to Weather.}
\label{tab:weather_variables}
\resizebox{\textwidth}{!}{%
\begin{tabular}{|c|c|c|c|}
\hline
\textbf{Weather Features }            & \textbf{Description}                                                   & \textbf{Unit}                   & \textbf{Range}     \\ \hline
Cloud Coverage               & Amount of Clouds in the Sky                                   & Okta                   & 0 to 9       \\
Air Pressure                 & Pressure within the atmosphere of the earth                   & Millibars              & 964 to 1043  \\
Temperature                  & Intensity of the heat present in the air                      & Celsius                & -24.9 to 32.5  \\
Relative Humidity            & Ratio of how much water vapor is in the air                   & Percentage             & 17 to 100     \\
Wind Direction               & The direction of the wind that is blowing at a given location & Degree                 & 1 to 360     \\
Wind Speed(Max)              & The maximum rate at which air is moving                       & Mile per second        & 0.5 to 19.3        \\
Wind Speed(Average)          & The average rate at which air is moving                       & Mile per second        & 0.2 to 8.6        \\
Precipitation                & Any product of the condensation of atmospheric water vapor    & Millimeter             & 0 to 19.6  \\
Global Horizontal Irradiance & The total solar radiation incident on a horizontal surface    & Watts per square meter & -1 to 886        \\
Sunshine                     & Direct sunlight duration without being covered by clouds      & Minutes                & 0 to 60     \\ \hline
\end{tabular}%
}
\end{table*}

\section{Data Processing} \label{dataprocessing}

This section presents data processing, including feature selection, outlier handling, and dependent variable transformation. Data analysis is performed by Python version 3.8.3. and the Jupyter Notebook version 6.0.3 was utilized as the main computing platform for editing and running codes. Moreover, a list of libraries used in the implementation is given in Table \ref{tab:library}.
Missing values were handled by linear interpolation, and the set of input features is extended. Three derived features, including \textit{Months}, \textit{Hours} and \textit{Prior Output Power} are added. These features are expected to play a major role in increasing the prediction accuracy when training the model. 

\begin{table}[h]
\centering
\captionsetup{width=.8\linewidth}
\caption{List of tools and libraries used in the implementation} \label{tab:library}
\begin{tabular}{|l|l|}
\hline
\textbf{Tool/Library name}  & \textbf{Version} \\ \hline
Python  & 3.8.3 \\
Jupyter Notebook & 6.0.3 \\
Pandas     & 1.0.5   \\
Numpy      & 1.19.5  \\
Sklearn    & 1.0.2   \\
Scipy      & 1.5.0   \\
Matplotlib & 3.2.2   \\
Seaborn    & 0.10.1  \\
XGBoost    & 1.5.0   \\ \hline
\end{tabular}
\end{table}

\subsection{Feature Selection}
ML involves selecting a subset of relevant features to be used in modeling. Therefore, redundant or irrelevant features and strongly correlated ones can be removed by a feature selection approach from a dataset without much loss of information \cite{lasso}. To strategically select the set of features, two methods, Pearson Correlation and Permutation-based Feature Importance, are employed.

\subsubsection{Pearson Correlation}
\begin{figure}[!h]
\centering
\includegraphics[scale = 0.69]{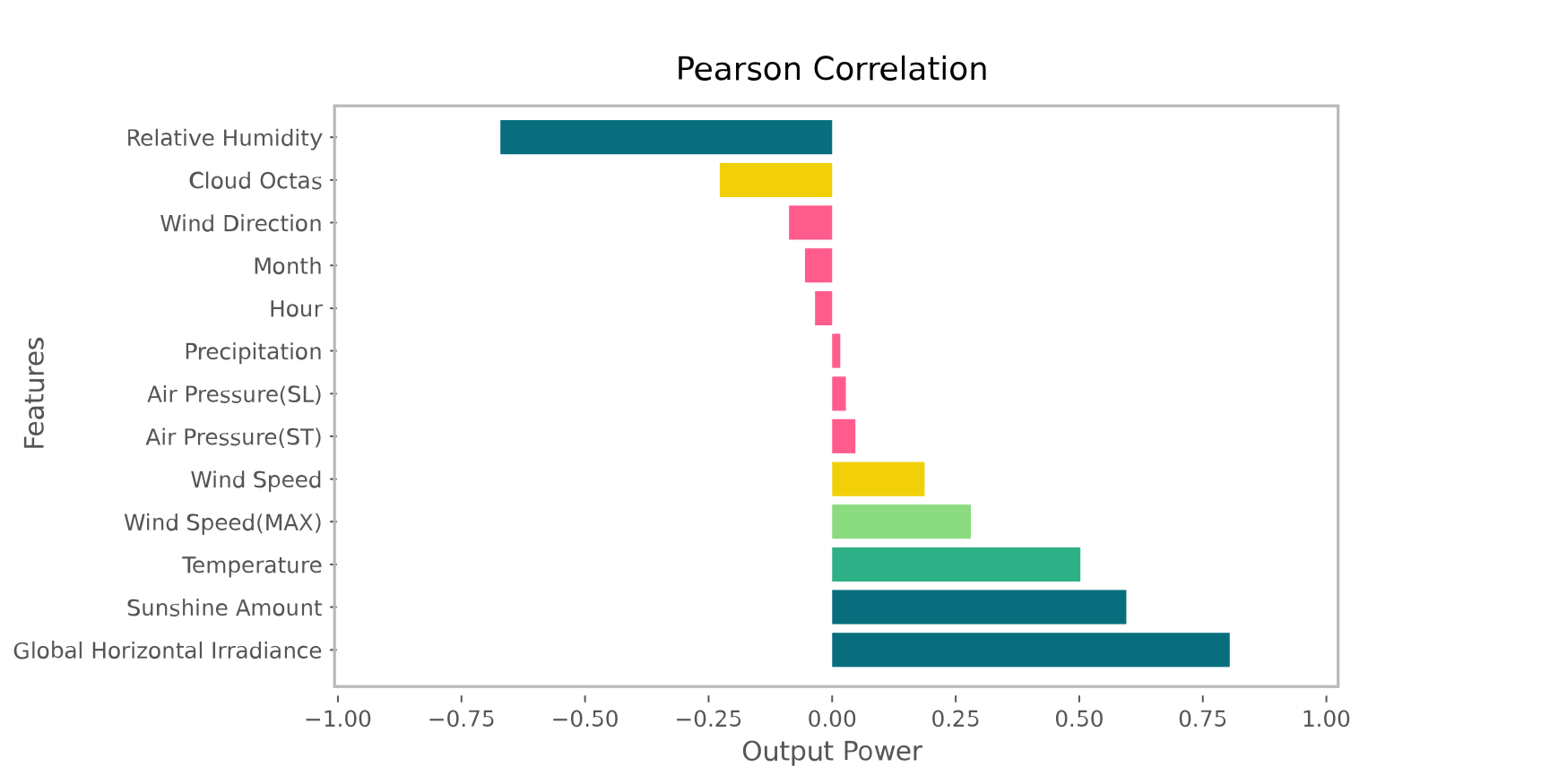}
\caption{Pearson Correlation}
\label{fig:cor}
\end{figure}

The Pearson correlation coefficient measures the linear relationship between two variables $X$ and $Y$ \cite{zhou, yu2019influence}. 
Accordingly, the linear relationship between PV output power and input features is illustrated in Figure \ref{fig:cor}. 

A Pearson correlation can range from strong to weak, or it can be zero. Therefore, a strong relationship is represented by 1, and a strong negative relationship is represented by -1. Zero means no relationship between two given features. 
As illustrated in Figure \ref{fig:cor}, the relationship between GHI and output power is strongly positive.
Furthermore, the output power relationship with sunshine amount and temperature is strongly positive. However, the relationship of output power with humidity and cloud coverage is strongly negative.

\subsubsection{Permutation-based Feature Importance}
\begin{figure}[!h]
\centering\includegraphics[scale = 0.69]{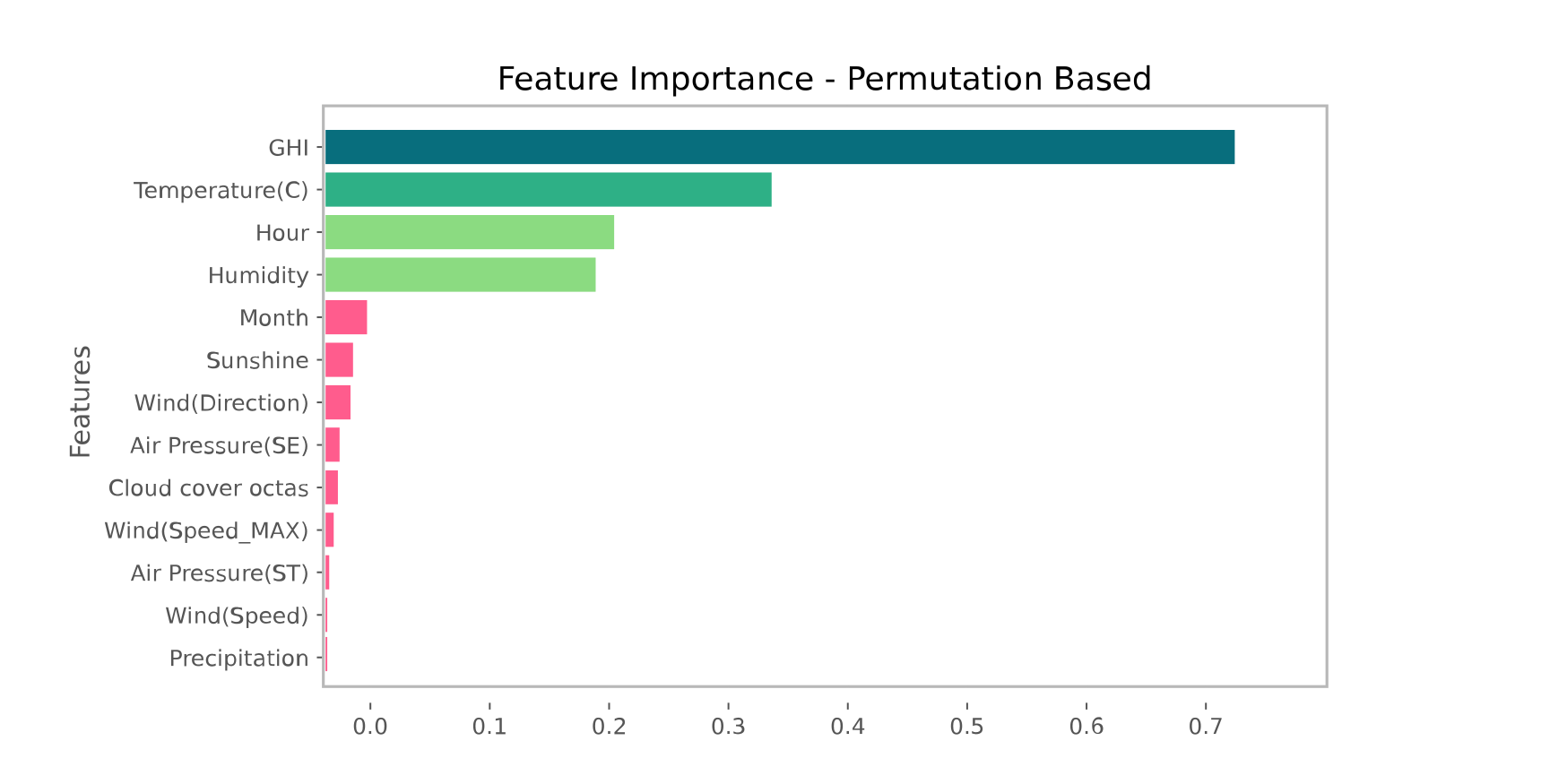}
\caption{Permutation-Based Feature Importance}
\label{fig:importance_permutation}
\end{figure}

Input features are used to calculate the feature importance score for a given model. Scores reflect the importance of each feature and indicate the impact of a specific feature on the final output prediction. 
Therefore, permutation-based feature importance is proposed in Figure \ref{fig:importance_permutation} to measure how influential each feature is in determining the splits. Permutation-based feature importance measures the decrease in overall performance when a single feature is removed.
In fact, the performance deterioration measures the importance of the removed variable. The idea is to measure how the feature has negatively affected the performance metric.


\subsection{Outlier Handling}

The term outlier refers to observations that appear to be inconsistent with the rest of the data \cite{barnet}. 
PV systems can suffer from varying issues that affect their output power, leading to anomalous values. Although some outliers are the result of natural variation, others are errors; thus, outlier detection is necessary to improve the model's accuracy. According to \cite{esc}, the term ``appears to be inconsistent" is the major challenge regarding outliers. Therefore, the outlier detection step in this study aims to address this challenge.

\begin{figure*}[!h]
    \centering\includegraphics[scale = 0.6]{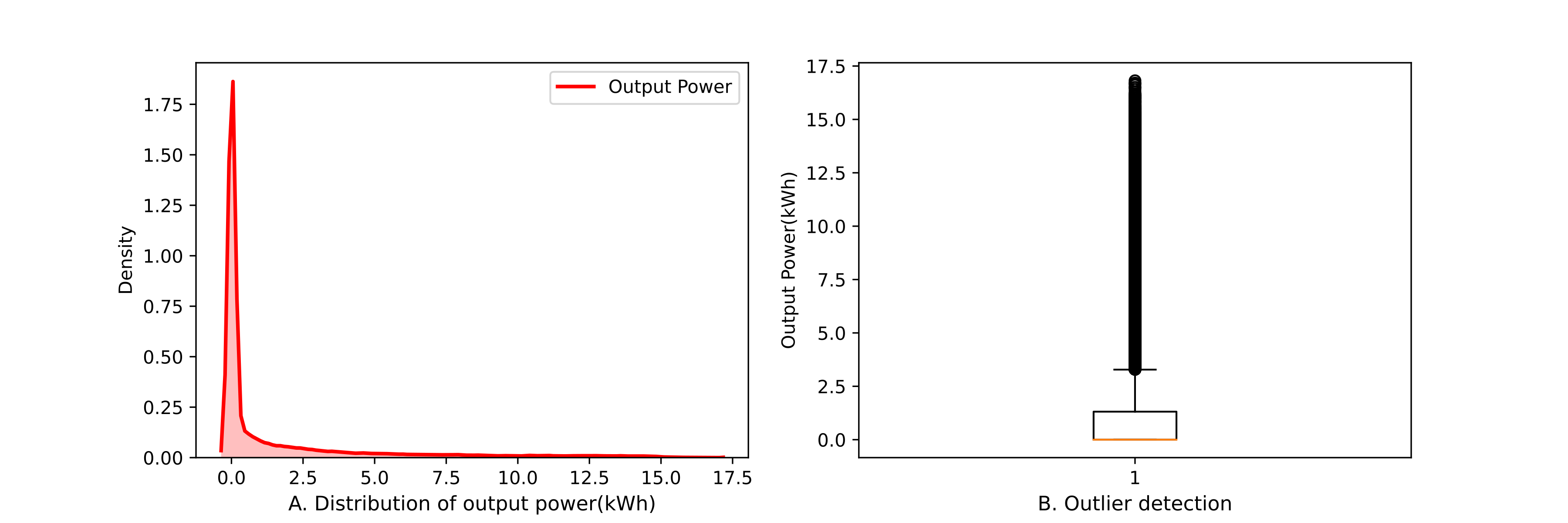}
    \caption{Actual Distribution(A) and Outlier Detection(B) of Output Power}
    \label{fig:withzero}
\end{figure*}

The Boxplot rule representation of outlier shows the best accuracy and robustness under PV errors \cite{lehman}, as shown in Figure \ref{fig:withzero}(B). 
After visually inspecting the solar data as shown in \ref{fig:withzero}(A), Interquartile Range (IQR) proximity rule was used. According to the IQR rule, values that fall outside boundaries are considered to be outliers based on the given Equation \ref{eq:ou} \cite{orille}.
\begin{equation}\label{eq:ou}
    \left \{\begin{array}{l}
    \text {Lowerboundary} =Q_{1}-1.5 IQR \\
    \text{Upperboundary} =Q_{3}+1.5 IQR \\
    \text{IQR} =Q_{3}-Q_{1} 
    \end{array}\right.
\end{equation}

\subsection{Target Transformation}
The distribution of the output power is highly right-skewed, as is illustrated in Figure \ref{fig:withzero}(A). Asymmetry is measured by skewness \cite{heyman} to explain the deviates from the normal distribution. To compute the skewness of the data, the Equation \ref{eq:skew} is given:  
\begin{equation}\label{eq:skew}
skewness = \frac{n}{(n-1)(n-2)} \sum_{x \in X}\left(\frac{x-\bar{x}}{\sigma}\right)^{3}
\end{equation}
The number of values, mean and standard deviation is denoted by $n$, $\bar{x}$ and $\sigma$ respectively.

\begin{table}[!h]
    \centering
    \begin{tabular}{|l|l|}
        \hline
         Skewness before removing zeros & 2.4 \\ \hline
         Skewness after removing zeros & 1.02 \\ \hline
         Skewness after applying transformation & 0.36 \\ \hline
    \end{tabular}
    \caption{Skewness of Output Power}
    \label{tbl:skew}
\end{table}

\begin{figure*}[!h]
\centering\includegraphics[scale = 0.6]{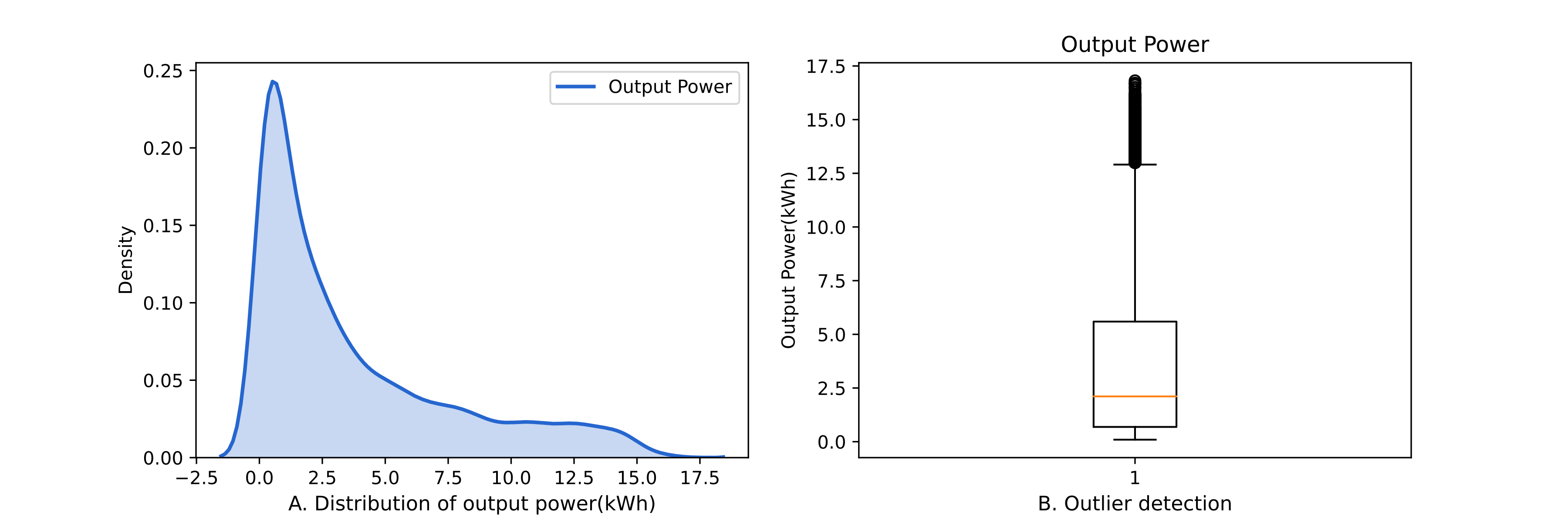}
\captionsetup{margin=.5cm}
\caption{Zeros are Excluded from the Distribution(A) and Outlier Detection(B) of Output Power}
\label{fig:withoutzeros}
\end{figure*}

\begin{figure*}[thpb]
\centering\includegraphics[scale = 0.6]{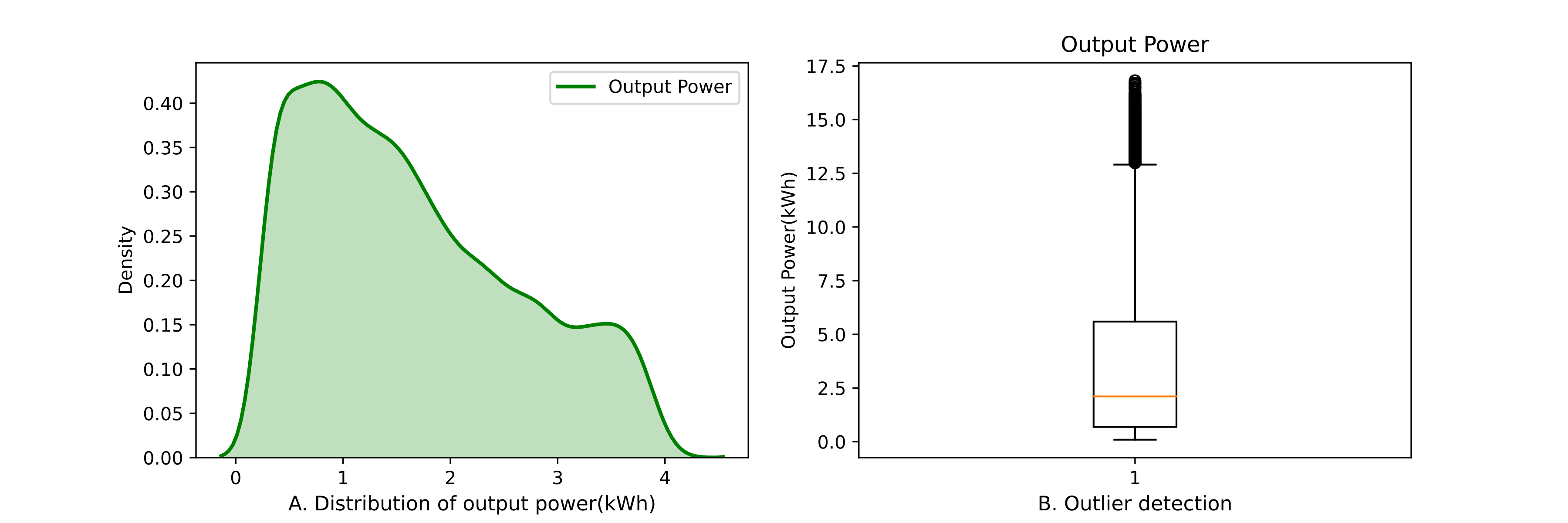}
\caption{Applying Transformation to the Output Power(Zeros are Excluded)}
\label{fig:training_after_transformation}
\end{figure*}

\begin{figure*}[thpb]
\centering\includegraphics[scale = 0.6]{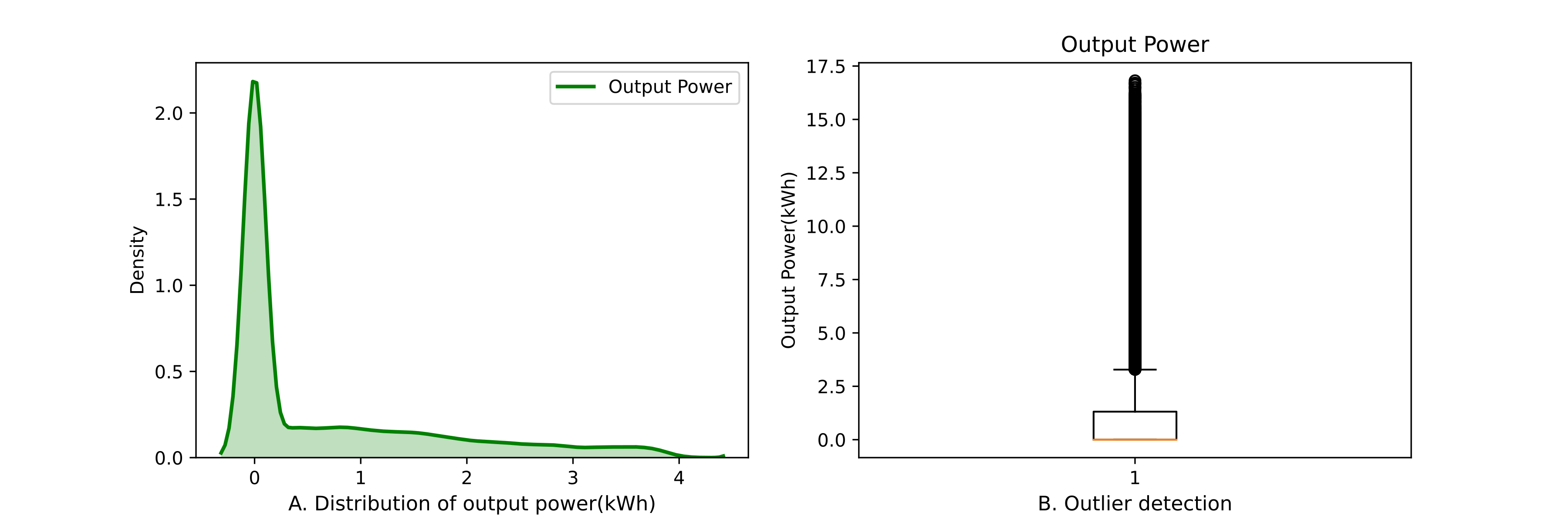}
\caption{Applying Transformation to the Output Power(Zeros are Included)}
\label{fig:sqrt}
\end{figure*}

Skewness was computed using the skew function available in SciPy library \cite{haung}. The skewness value is represented in Table \ref{tbl:skew}. 
If the skewness value falls above or below +1 or -1, the data is highly right or left-skewed, respectively. The data has a moderately skewed distribution when the computed skewness lies between -1 and -0.5 or between +0.5 and 1. Additionally, if the skewness is between -0.5 and 0.5, the data distribution is nearly symmetrical. Finally, when the computed skewness is equal to zero, then the data is symmetric \cite{perrot}. Table \ref{tbl:skew} represents the skewness of the output power. PV output power is depended on how much light they receive. Zeros are usually the result of nighttime data assigned to the PV system's output. Therefore, all zeros are temporarily excluded from the output power variable as shown in Figure \ref{fig:withoutzeros} to detect outliers in actual data. However, nighttime data is included in the final predictions to use the whole day to identify time regions. Therefore, as illustrated in Figure \ref{fig:training_after_transformation}, square root transformation is applied to the output power to normalize the data distribution.

After inspecting the source of outliers, zeros are included again, and the transformation is applied to PV output power as is shown in Figure \ref{fig:sqrt}.
Therefore, depending upon the data, taking square root transformation can probably stabilize the variance of the distribution by decreasing the skewness. Moreover, extreme values are penalized more by transformation.

\section{Implementation} \label{implementation}

In this section, implementation details are presented regarding different ML models. 
Standardization is performed for input features 
to ensure that features with wider ranges do not dominate the distance metric. Moreover, square root transformation is only applied to the target variable. In general, the train-test split is carried out with an 80:20 ratio. The best set of hyperparameters for a given ML model is tuned by random search. The implemented code, along with collected data, can be accessed from the public Github Repository at \cite{data_githubrepo}. 

\subsection{XGBoost}
Random search is adopted to extract optimal values for hyperparameters of XGBoost as is given in Table \ref{tab:xgboost_hyper}.
The number of trees is defined by \(n\_estimator\). In general, the higher the number of trees, the better model learns from the data. 
Thus, trees should be added until there is no further improvement in the model since the model overfits as more and more trees are added \cite{Kumar}. learning\_rate determines step size shrinkage to update the weight to prevent overfitting.

Additionally, the fraction of observations to be randomly sampled is denoted by subsample.
Moreover, \(max\_depth\) controls overfitting. Increasing \(max\_depth\) will make the model more complex. Hence, the model is more likely to overfit. 
Accordingly, the best accuracy is achieved with the hyperparameter values given in Table \ref{tab:xgboost_hyper}.

\begin{table}[H]
\centering
\caption{XGBoost Hyperparameters and their values}
\label{tab:xgboost_hyper}
\scalebox{0.40}{
\resizebox{\textwidth}{!}{%
\begin{tabular}{c|ccc|}
\cline{2-4}
                                      & \multicolumn{3}{c|}{\textbf{Values}}                                                              \\ \hline
\multicolumn{1}{|c|}{\textbf{Hyperparameters}} & \multicolumn{1}{c|}{30 Min} & \multicolumn{1}{l|}{1 Hour} & \multicolumn{1}{l|}{4 Hours} \\ \hline
\multicolumn{1}{|c|}{n\_estimator}    & \multicolumn{1}{c|}{450}    & \multicolumn{1}{c|}{300}    & 250                          \\ \hline
\multicolumn{1}{|c|}{learning\_rate}  & \multicolumn{1}{c|}{0.1}    & \multicolumn{1}{c|}{0.1}    & 0.1                          \\ \hline
\multicolumn{1}{|c|}{subsample}       & \multicolumn{1}{c|}{0.6}    & \multicolumn{1}{c|}{0.6}    & 0.6                          \\ \hline
\multicolumn{1}{|c|}{max\_depth}      & \multicolumn{1}{c|}{10}     & \multicolumn{1}{c|}{5}      & 5                            \\ \hline
\multicolumn{1}{|c|}{gamma}           & \multicolumn{1}{c|}{0.1}    & \multicolumn{1}{c|}{0.1}    & 0.1                          \\ \hline
\end{tabular}%
}}
\end{table}

\subsection{Random Forest}
Random Forest has several hyperparameters to be set. Performance achieved by some default hyperparameters was optimal. However, two of the hyperparameters are required to be set. \(n\_estimator\) is the number of trees in the forest, and \(max\_features\) determines the maximum number of allowed features to be used in individual trees.

\begin{table}[H]
\centering
\caption{RF Hyperparameters and their values}
\label{tab:rf_hyper}
\scalebox{0.40}{
\resizebox{\textwidth}{!}{%
\begin{tabular}{c|ccc|}
\cline{2-4}
                                      & \multicolumn{3}{c|}{\textbf{Values}}                                                              \\ \hline
\multicolumn{1}{|c|}{\textbf{Hyperparameters}} & \multicolumn{1}{c|}{30 Min} & \multicolumn{1}{l|}{1 Hour} & \multicolumn{1}{l|}{4 Hours} \\ \hline
\multicolumn{1}{|c|}{n-estimator}     & \multicolumn{1}{c|}{350}    & \multicolumn{1}{c|}{300}    & 200                          \\ \hline
\multicolumn{1}{|c|}{max\_features}   & \multicolumn{1}{c|}{log2}   & \multicolumn{1}{c|}{log2}   & log2                         \\ \hline
\multicolumn{1}{|c|}{max\_depth}      & \multicolumn{1}{c|}{None}   & \multicolumn{1}{c|}{None}   & None                         \\ \hline
\end{tabular}%
}}
\end{table}

\subsection{K-Nearest Neighbour}
The most crucial hyperparameter to be set is the number of neighbors denoted by \(n\_neighbour\). A range of values was assigned to the \(n\_neighbour\), and finally, the optimal one is given in Table \ref{tab:knn_hyper}.

\begin{table}[H]
\centering
\caption{KNN Hyperparameters and their values}
\label{tab:knn_hyper}
\scalebox{0.40}{
\resizebox{\textwidth}{!}{%
\begin{tabular}{c|ccc|}
\cline{2-4}
                                      & \multicolumn{3}{c|}{\textbf{Values}}                                                                \\ \hline
\multicolumn{1}{|c|}{\textbf{Hyperparameters}} & \multicolumn{1}{c|}{30 Min}  & \multicolumn{1}{l|}{1 Hour}  & \multicolumn{1}{l|}{4 Hours} \\ \hline
\multicolumn{1}{|c|}{n-neighbours}    & \multicolumn{1}{c|}{3}       & \multicolumn{1}{c|}{3}       & 3                            \\ \hline
\multicolumn{1}{|c|}{p}               & \multicolumn{1}{c|}{1}       & \multicolumn{1}{c|}{1}       & 1                            \\ \hline
\multicolumn{1}{|c|}{weights}         & \multicolumn{1}{c|}{uniform} & \multicolumn{1}{c|}{uniform} & uniform                      \\ \hline
\end{tabular}%
}}
\end{table}

\subsection{Multilayer Perceptron}

The values of the hyperparameters are given in Table \ref{tab:mlp_hyper}. The hidden layer size was assigned by random search. Two hidden layers, each containing eighty-five and sixty-five neurons, respectively, are set to be optimal values. Additionally, \(max\_iter\) was set to determine the maximum number of epochs in which the model trains the data.
The default values were set for the rest of the hyperparameters.

\begin{table}[H]
\centering
\caption{MLP Hyperparameters and their values}
\label{tab:mlp_hyper}
\scalebox{0.42}{
\resizebox{\textwidth}{!}{%
\begin{tabular}{c|ccc|}
\cline{2-4}
                                          & \multicolumn{3}{c|}{\textbf{Values}}                                                                         \\ \hline
\multicolumn{1}{|c|}{\textbf{Hyperparameters}}     & \multicolumn{1}{c|}{30 Min}        & \multicolumn{1}{l|}{1 Hour}     & \multicolumn{1}{l|}{4 Hours} \\ \hline
\multicolumn{1}{|c|}{hidden\_layer\_size} & \multicolumn{1}{c|}{(80,80,80,80)} & \multicolumn{1}{c|}{(80,80)}    & (80,80)                      \\ \hline
\multicolumn{1}{|c|}{max\_iter}           & \multicolumn{1}{c|}{200}           & \multicolumn{1}{c|}{200}        & 200                          \\ \hline
\multicolumn{1}{|c|}{activation}          & \multicolumn{1}{c|}{relu}          & \multicolumn{1}{c|}{relu}       & relu                         \\ \hline
\multicolumn{1}{|c|}{solver}              & \multicolumn{1}{c|}{adam}          & \multicolumn{1}{c|}{adam}       & adam                         \\ \hline
\multicolumn{1}{|c|}{learning\_rate}      & \multicolumn{1}{c|}{invscaling}    & \multicolumn{1}{c|}{invscaling} & invscaling                   \\ \hline
\multicolumn{1}{|c|}{batch\_size}         & \multicolumn{1}{c|}{auto}          & \multicolumn{1}{c|}{auto}       & auto                         \\ \hline
\multicolumn{1}{|c|}{alpha}               & \multicolumn{1}{c|}{0.0001}        & \multicolumn{1}{c|}{0.0001}     & 0.0001                       \\ \hline
\end{tabular}%
}}
\end{table}

\subsection{Support Vector Regression}
Apart from all default values given to hyperparameters as optimal ones, \textit{Gamma} and $C$ were set by different values as given in Table \ref{tab:svr_hyper}.

Kernel function transforms a low-dimensional input space into a higher-dimensional feature space. RBF is used when the dataset is not linearly separable. Furthermore, \textit{Gamma} is a scaling parameter to determine the spread of the kernel. A large \textit{Gamma} results in a narrow kernel; thus, the local influence of each single data point increases when \textit{Gamma} is larger. Moreover, $C$ represents the regularization parameter to control the errors and avoid overfitting. The lower the value of $C$, the larger margin of the decision boundary is chosen and vice versa. 


\begin{table}[H]
\centering
\caption{SVR Hyperparameters and their values}
\label{tab:svr_hyper}
\scalebox{0.40}{
\resizebox{\textwidth}{!}{%
\begin{tabular}{c|ccc|}
\cline{2-4}
                                      & \multicolumn{3}{c|}{\textbf{Values}}                                                              \\ \hline
\multicolumn{1}{|c|}{\textbf{Hyperparameters}} & \multicolumn{1}{c|}{30 Min} & \multicolumn{1}{l|}{1 Hour} & \multicolumn{1}{l|}{4 Hours} \\ \hline
\multicolumn{1}{|c|}{kernel}          & \multicolumn{1}{c|}{rbf}    & \multicolumn{1}{c|}{rbf}    & rbf                          \\ \hline
\multicolumn{1}{|c|}{gamma}           & \multicolumn{1}{c|}{0.8}    & \multicolumn{1}{c|}{0.8}    & 0.6                          \\ \hline
\multicolumn{1}{|c|}{C}               & \multicolumn{1}{c|}{4}      & \multicolumn{1}{c|}{4}      & 5                            \\ \hline
\multicolumn{1}{|c|}{epsilon}         & \multicolumn{1}{c|}{0.1}    & \multicolumn{1}{c|}{0.1}    & 0.1                          \\ \hline
\end{tabular}%
}}
\end{table}

\subsection{Performance Metrics}

Evaluation of the ML models is a key aspect of measuring the prediction performance. The percentage of correct predictions determines the accuracy of a model.
Although there are various performance metrics to be used, there is no general consensus on a set of acceptable ones \cite{damer}, \cite{khalid2020survey} 
However, widely used metrics in PV output power are proposed for evaluating the model performance, such as MAE, MAPE, RMSE \cite{theocharides} and R-squared.
Different results are expected because performance evaluation metrics are defined differently.

\begin{figure*}[!t]
\centering
\centering\includegraphics[scale = 1.2]{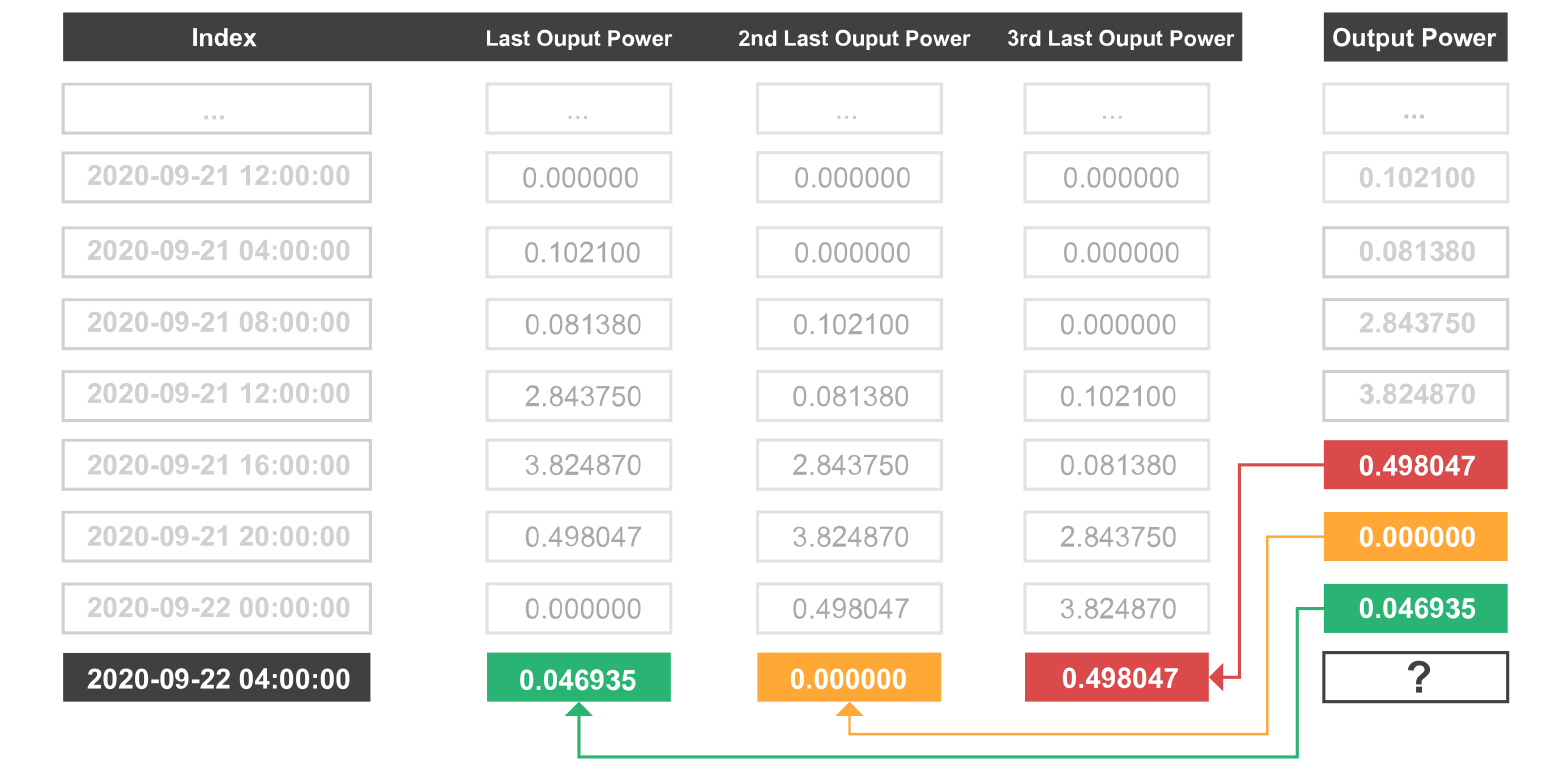}
\caption{A set of prior output power as input features}
\label{fig:guidance}
\end{figure*}

\section{Importance of Prior Output Power as the Derived Feature}\label{prior}

Initially, along with weather-related features, two derived features, including ``Months" and ``Hours," were employed as the input features. While having a model with high accuracy is essential, the key purpose of this research was to improve the accuracy in an efficient way. Therefore, another derived feature was used during the training of the data, and the set of input features was extended again. As a result, the accuracy increased almost noticeably after adding one prior output power as the input feature. For instance, ``No Prior" represents currently measured actual output power, ``1 Prior" indicates the last measured, ``2 Prior" and ``3 Prior" indicates two and three last measured output power in a given time frame, as is illustrated in Figure \ref{fig:guidance}. 

\begin{figure}[h]
\centering\includegraphics[scale = 0.55]{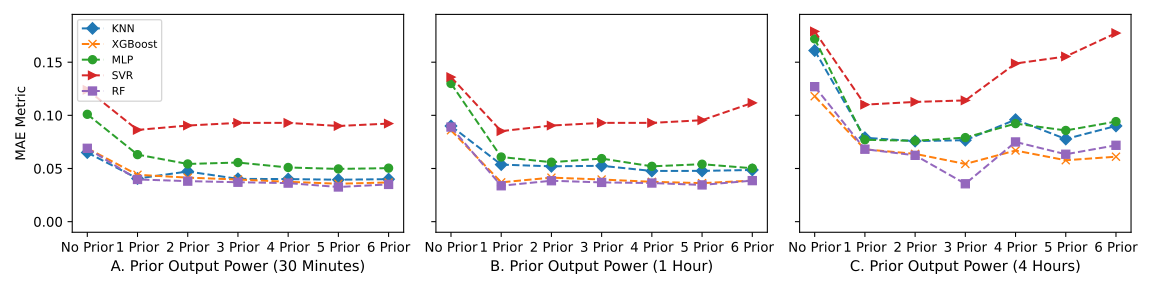}
\caption{Measured accuracy for different sets of prior output power (MAE)}
\label{fig:pr_1}
\end{figure}

\begin{figure}[h]
\centering
\centering\includegraphics[scale = 0.55]{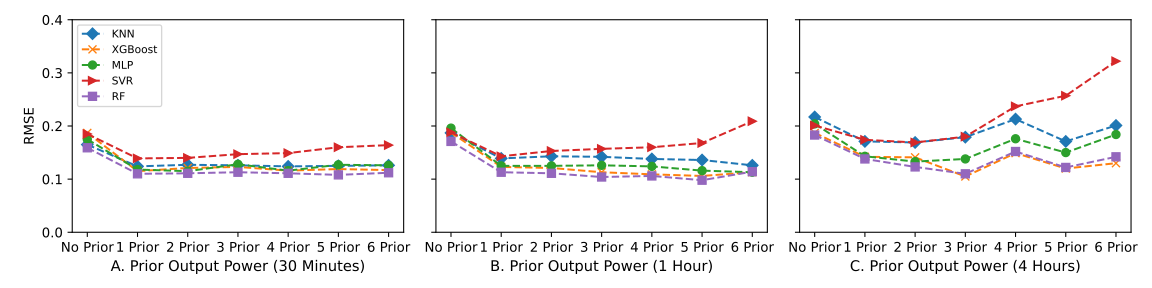}
\caption{Measured accuracy for different sets of prior output power (RMSE)}
\label{fig:pr_2}
\end{figure}

\begin{figure}[h]
\centering
\centering\includegraphics[scale = 0.55]{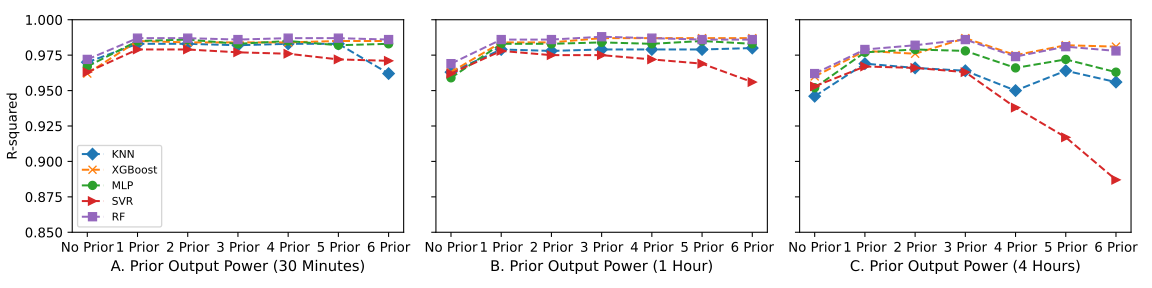}
\caption{Measured accuracy for different sets of prior output power (R-squared)}
\label{fig:pr_3}
\end{figure}

\begin{figure}[h]
\centering
\centering\includegraphics[scale = 0.55]{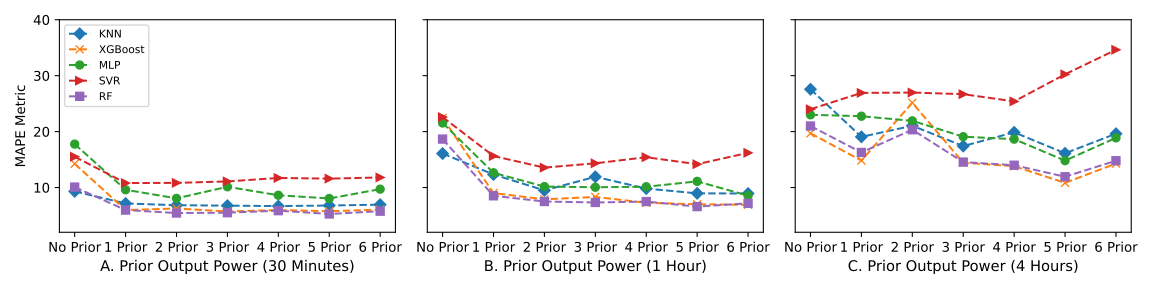}
\caption{Measured accuracy for different sets of prior output power (MAPE)}
\label{fig:pr_4}
\end{figure}

Figures \ref{fig:pr_1} - \ref{fig:pr_4}, illustrate how the accuracy is changed when a set of prior output power is added as input features. The X-axis represents the number of previously measured output power as input features in three different time frames, and the Y-axis represents the performance metric of the measured output power. Initially, the first obtained accuracy on the X-axis (No Prior) is obtained without adding prior output powers as input features. 

Adding one prior output power confirms a good accuracy of the applied approach for all models compared to what was measured initially without adding prior output power. Furthermore, a constant level of accuracy is observed by adding more than one prior output power as the input feature, except for SVR. Accordingly, by taking each model's differences into account, SVR has drawn attention since a steady and sharp decrease in obtained accuracy is noticed in one and 4 hours intervals, respectively.

However, SVR has drawn attention noticeably since a very sharp decrease in obtained accuracy is observed after adding four prior output power as the input feature. Moreover, all models have shown some fluctuations in obtained accuracy after adding three prior output power. Additionally, all results are given in Appendix \ref{appendix:a} as the supplementary file.

\section{Results and Discussion} \label{results}
This section provides accuracy obtained by different ML models as well as CV scores for each given ML model. Moreover, the overall discussion is centered around the findings and challenges.
\begin{figure*}[!ht]
\centering\includegraphics[scale = 0.60]{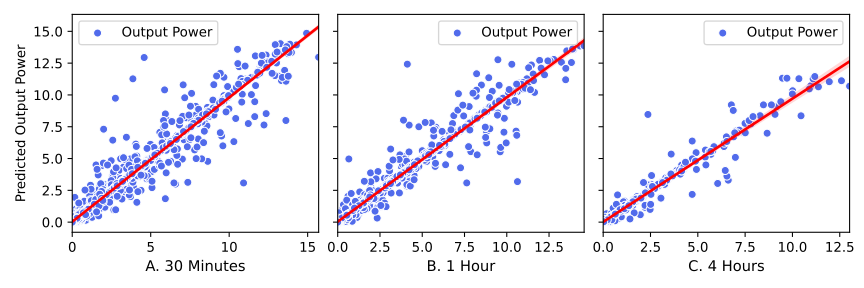}
\caption{Predicted (Y-axis) Versus Actual (X-axis) Output Power(kWh) for KNN}
\label{fig:knn_fig}
\end{figure*}


\subsection{Accuracy Results}
\begin{figure*}[t]
\centering\includegraphics[scale = 0.60]{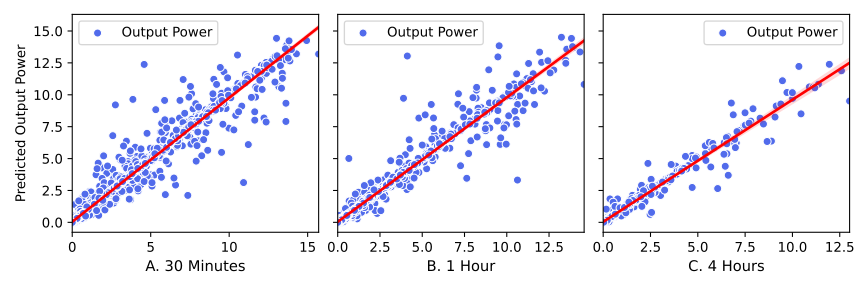}
\caption{Predicted (Y-axis) Versus Actual (X-axis) Output Power(kWh) for XGBoost}
\label{fig:XGboost_result}
\end{figure*}

\begin{figure*}[t]
\centering\includegraphics[scale = 0.60]{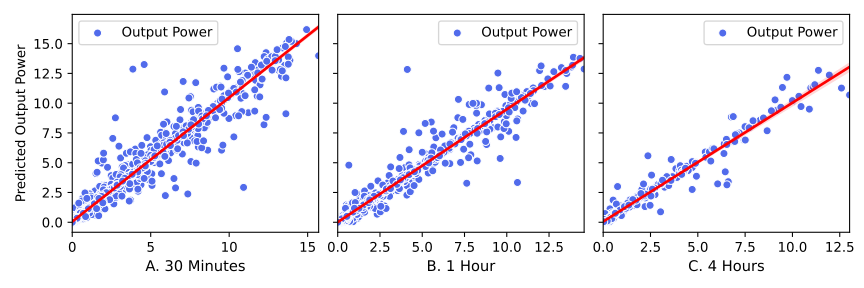}
\caption{Predicted (Y-axis) Versus Actual (X-axis) Output Power(kWh) for MLP}
\label{fig:mlp_figure}
\end{figure*}

\begin{figure*}[t]

\centering
\centering\includegraphics[scale = 0.60]{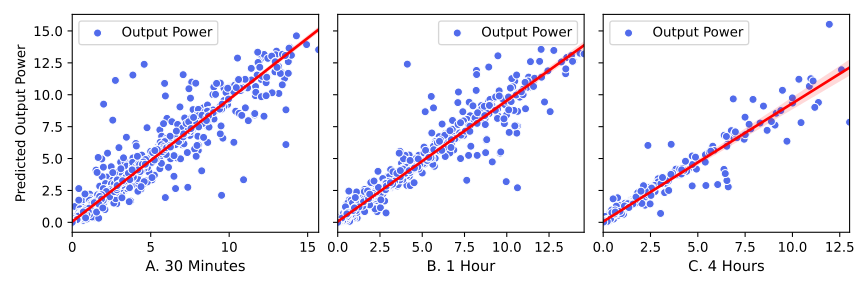}
\caption{Predicted (Y-axis) versus Actual (X-axis) Output Power(kWh) for SVR}
\label{fig:svr_figure}
\end{figure*}

\begin{figure*}[t]

\centering
\centering\includegraphics[scale = 0.60]{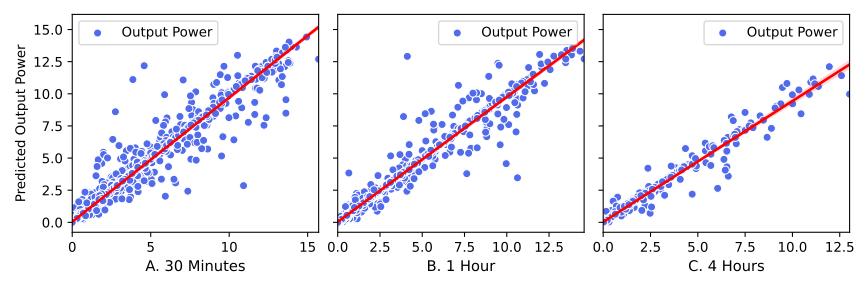}
\caption{Predicted (Y-axis) versus Actual (X-axis) Output Power(kWh) for RF}
\label{fig:RF_figure}
\end{figure*}

Figure \ref{fig:knn_fig} - \ref{fig:RF_figure} 
represent the actual (X-axis) and predicted data points (Y-axis) in case of KNN, XGBoost, MLP, SVR, and RF.
The red line represents a perfect prediction (referring to the closeness of the predicted value with that of the actual value in the dataset) in the scatter plot. The red line indicates how well the predicted values match the actual data point.
Furthermore, the number of input features is reduced. GHI, temperature, humidity, months, hours, and one prior output power are the main features to include in the predictions. 

By visually inspecting the figures and according to the best fit line, the difference in the values of the predicted and actual data is quite not much in all time frames with a similar prediction pattern in all time frames. However, very few poorly predicted data points are observed mainly in 30 minutes time frame. Concerning tables, although the accuracy tends to decrease slightly as the time frame increases, MAPE outlines a rapid decrease in accuracy. To calculate MAPE, zeros are excluded since MAPE takes undefined values when actual data points are zero. Moreover, as the time frame increases, the data size decreases, and accuracy tends to decrease. All in all, the figures outline a very good accuracy regarding obtained results. 
Table \ref{tbl:all_results} presents the performance metric results. Moreover, the actual and predicted standard deviation of the test set is given in Table \ref{tbl:std_results_all} in which the difference between the set of actual and predicted values is small.



\begin{table*}
\centering
\caption{Accuracy results for each given ML model}
\label{tbl:all_results}
\def\arraystretch{1.2}
\resizebox{\textwidth}{!}{%
\label{tab:my-table}
\begin{tabular}{|c|c|c|c|c|c|c|c|c|c|c|c|c|c|c|c|} 
\hline
 & \multicolumn{3}{c|}{\textbf{KNN}} & \multicolumn{3}{c|}{\textbf{XGBoost}} & \multicolumn{3}{c|}{\textbf{MLP}} & \multicolumn{3}{c|}{\textbf{SVR}} & \multicolumn{3}{c|}{\textbf{RF}} \\ 
\cline{2-16}
\textbf{Metrics} & \textbf{30 Min} & \textbf{1 Hour} & \textbf{4 Hours} & \textbf{30 Min} & \textbf{1 Hour} & \textbf{4 Hour} & \textbf{30 Min} & \textbf{1 Hour} & \textbf{4 Hour} & \textbf{30 Min} & \textbf{1 Hour} & \textbf{4 Hour} & \textbf{30 Min} & \textbf{1 Hour} & \textbf{4 Hours} \\ 
\hline
\textbf{MAPE} & 7.15 & 12.33 & 19.06 & 5.97 & 9.01 & 14.83 & 9.60 & 12.63 & 22.75 & 10.83 & 15.63 & 26.92 & 5.27 & 8.52 & 16.27 \\ 
\hline
\textbf{MAE} & 0.040 & 0.053 & 0.078 & 0.036 & 0.044 & 0.068 & 0.057 & 0.063 & 0.077 & 0.080 & 0.086 & 0.110 & 0.032 & 0.039 & 0.068 \\ 
\hline
\textbf{RMSE} & 0.125 & 0.139 & 0.170 & 0.115 & 0.123 & 0.142 & 0.118 & 0.126 & 0.147 & 0.139 & 0.143 & 0.174 & 0.110 & 0.113 & 0.138 \\ 
\hline
\textbf{R2} & 0.983 & 0.979 & 0.964 & 0.981 & 0.984 & 0.978 & 0.985 & 0.983 & 0.977 & 0.980 & 0.978 & 0.967 & 0.987 & 0.986 & 0.979 \\
\hline
\end{tabular}%
}
\end{table*}

\begin{table*}
\centering
\caption{Standard Deviation of Actual and Predicted Test Set}
\label{tbl:std_results_all}
\def\arraystretch{1.2}
\resizebox{\textwidth}{!}{%
\begin{tabular}{|c|c|c|c|c|c|c|c|c|c|c|c|c|c|c|c|} 
\hline
\multirow{2}{*}{\begin{tabular}[c]{@{}c@{}}\textbf{STD}\\\textbf{ Deviation}\end{tabular}} & \multicolumn{3}{c|}{\textbf{KNN}} & \multicolumn{3}{c|}{\textbf{XGBoost}} & \multicolumn{3}{c|}{\textbf{MLP}} & \multicolumn{3}{c|}{\textbf{SVR}} & \multicolumn{3}{c|}{\textbf{RF}} \\ 
\cline{2-16}
 & \textbf{30 Min} & \textbf{1 Hour} & \textbf{4 Hours} & \textbf{30 Min} & \textbf{1 Hour} & \textbf{4 Hour} & \textbf{30 Min} & \textbf{1 Hour} & \textbf{4 Hour} & \textbf{30 Min} & \textbf{1 Hour} & \textbf{4 Hour} & \textbf{30 Min} & \textbf{1 Hour} & \textbf{4 Hours} \\ 
\hline
\textbf{Test Set} & 0.973 & 0.982 & 0.971 & 0.973 & 0.982 & 0.971 & 0.973 & 0.982 & 0.971 & 0.973 & 0.982 & 0.971 & 0.973 & 0.982 & 0.971 \\ 
\hline
\textbf{Predicted} & 0.970 & 0.984 & 0.964 & 0.971 & 0.980 & 0.957 & 0.960 & 0.990 & 0.966 & 0.961 & 0.970 & 0.932 & 0.951 & 0.977 & 0.942 \\
\hline
\end{tabular}%
}
\end{table*}

\subsection{Cross-Validation Scores}
A ten-fold cross-validation is examined. Table \ref{tbl:rf_results} presents average of scores for each given ML model in three different time frames. 
In each fold, the performance metric is computed to determine the average performance. Compared to the accuracy obtained by the test set, results from CV demonstrate good prediction accuracy following ten-fold cross-validation.

\begin{table}[h]
\centering
\caption{Cross validation Scores}
  \label{tbl:rf_results}
\begin{tabular}{lcllcllcll}
        & \multicolumn{3}{c}{30 Min} & \multicolumn{3}{c}{Hourly} & \multicolumn{3}{c}{4 Hours} \\ \hline
KNN     & \multicolumn{3}{c}{0.983}  & \multicolumn{3}{c}{0.980}  & \multicolumn{3}{c}{0.968}  \\ \hline
XGBoost & \multicolumn{3}{c}{0.986}  & \multicolumn{3}{c}{0.981}  & \multicolumn{3}{c}{0.980}  \\ \hline
MLP      & \multicolumn{3}{c}{0.986}  & \multicolumn{3}{c}{0.984}  & \multicolumn{3}{c}{0.971}  \\ \hline
SVR     & \multicolumn{3}{c}{0.981}  & \multicolumn{3}{c}{0.977}  & \multicolumn{3}{c}{0.961}  \\ \hline
RF     & \multicolumn{3}{c}{0.988}   & \multicolumn{3}{c}{0.987}  & \multicolumn{3}{c}{0.980}  \\ \hline
\end{tabular}
\end{table}

\subsection{Overall Discussion}
Regarding methodology, there are certain aspects to improve the results. One approach is to split the data into different weather conditions \cite{khademi}. However, the proposed method is highly location-based since seasonal variation might not result in weather conditions. For instance, although more sunshine is expected during summer, but the wettest season is in summer in our case study. Therefore less sunshine is expected during summer, which makes the summer meteorological pattern more similar to fall and winter regarding sunshine hours.

Moreover, this research presents intraday predictions, and nighttime data is included to use the whole day to identify time regions. However, more reliability is expected by excluding zeros in the day-ahead or weak-ahead prediction. In that case, removing zeros does not harm our model because of missing time regions.

Furthermore, this study uses one prior output power as an input feature to predict the next output power. Therefore, using the currently measured actual output power as an input feature can be challenging in very short-time predictions because the model will regularly update itself. Accordingly, this challenge is a part of our future study directions.

\section{Conclusion}\label{conclusion}

This paper addresses the problem of predicting green energy availability in a smart building scenario based on real weather conditions. 
The main purpose of the proposal is to predict the energy availability in the next time frame (30 minutes, 1 hour, and 4 hours). Compared to state-of-the-art proposals, our used methodology suggests an abstraction layer focused on data processing and output production that can be tailored for different locations. The proposed solution captures the historical actual energy production from the roof-top PV panels installed in the university building in Estonia and the weather data from the nearby weather station. Five different ML models: KNN, XGBoost, RF, SVR, and MLP, were implemented and compared the prediction accuracy. Three different time frames (30 minutes, 1 hour, and 4 hours) were selected to fit and predict the data. Several metrics, such as MAPE, MAE, RMSE, and R2, are considered while comparing the performance of all the ML models. 
Compared to research studies in the field of PV predictions, our results show that the obtained prediction accuracy outperforms with performance gains reaching up to 14\% (R2) in case of the RF model and 25\% (MAPE) in case of the SVR model. The comparison result infers a similar performance when RF and XGBoost were applied. On the other hand, the ML models MLP, SVR, and KNN give a similar prediction pattern. However, when the time frame increases from 30 minutes to 4 hours, SVR tends to produce less prediction accuracy. From the extensive comparison, it can be concluded that tree-based algorithms (RF and XGBoost) are relatively producing better prediction accuracy and stability in all time frames.

The proposed model does rely on the historical dataset and makes the energy consumption forecasting based on a portion of the historical dataset. However, it does not update itself based on the current behavior of solar panels and the weather. Making the model online is a part of our future direction to work, where the model would constantly update itself with the predicted and actual values. Further, gathering real data from the different geographical locations would definitely improve the accuracy of the model. Another direction to work on is its implementation in a city-wide solar panel system. Currently, we are in the process of collaboration with Tartu City to gather the solar panel data and extend the prediction model. 


\section*{Acknowledgment}
This work is partially funded by the Ministry of Science and Technology (MOST) of Taiwan under grant number 111-2221-E-167-016.

\section*{Data Availability}
The data used in this manuscript can be obtained from \url{https://github.com/chinmaya-dehury/Loc_Green_Energy_Availability_Pred}\cite{data_githubrepo} Github repository. This public repository contains both the required data in .csv and in .xlsx formats and the python code in Jupyter notebook format. The data include the historical performance of solar panels, weather data, and global horizontal irradiance. 

\newpage
\bibliographystyle{elsarticle-num}
\bibliography{references}

\onecolumn
\appendix

\section{Tables for section 6}\label{appendix:a}

\begin{table}[H]
\centering
\caption{KNN results}
\label{tab:knn_prior}
\def\arraystretch{1}
\resizebox{\textwidth}{!}{%
\begin{tabular}{|c|ccc|ccc|ccc|ccc|ccc|ccc|} 
\cline{2-19}
\multicolumn{1}{c|}{} & \multicolumn{3}{c|}{\textbf{1 Prior}} & \multicolumn{3}{c|}{\textbf{2 Prior}} & \multicolumn{3}{c|}{\textbf{3 Prior}} & \multicolumn{3}{c|}{\textbf{4 Prior}} & \multicolumn{3}{c|}{\textbf{5 Prior}} & \multicolumn{3}{c|}{\textbf{6 Prior}}  \\ 
\hline
\textbf{Metrics}      & 30 Min & 1 Hour & 4 Hour              & 30 min & 1 Hour & 4 Hour              & 30 Min & 1 Hour & 4 Hour              & 30 min & 1 Hour & 4 Hour              & 30 min & 1 Hour & 4 Hour              & 30 min & 1 Hour & 4 Hour               \\ 
\hline
Mape                  & 7.15   & 12.33  & 19.06               & 6.84   & 9.49   & 21.04               & 6.76   & 11.92  & 17.41               & 6.68   & 9.81   & 19.87               & 6.78   & 8.96   & 16.12               & 6.94   & 8.93   & 19.62                \\
MAE                   & 0.040  & 0.053  & 0.078               & 0.040  & 0.05   & 0.075               & 0.040  & 0.052  & 0.076               & 0.040  & 0.047  & 0.095               & 0.039  & 0.047  & 0.077               & 0.039  & 0.048  & 0.090                \\
RMSE                  & 0.124  & 0.139  & 0.170               & 0.127  & 0.143  & 0.169               & 0.127  & 0.142  & 0.179               & 0.124  & 0.138  & 0.213               & 0.125  & 0.136  & 0.170               & 0.126  & 0.140  & 0.201                \\
R-Square              & 0.983  & 0.979  & 0.969               & 0.983  & 0.978  & 0.966               & 0.982  & 0.979  & 0.964               & 0.983  & 0.979  & 0.950               & 0.983  & 0.979  & 0.964               & 0.962  & 0.980  & 0.956                \\
STD of  Predicted     & 0.970  & 0.982  & 0.964               & 0.974  & 0.976  & 0.934               & 0.965  & 0.993  & 0.940               & 0.969  & 0.958  & 0.947               & 0.965  & 0.956  & 0.885               & 0.961  & 1.00   & 0.935                \\
STD of Test Set       & 0.973  & 0.984  & 0.971               & 0.982  & 0.980  & 0.929               & 0.976  & 1.00   & 0.948               & 0.975  & 0.962  & 0.955               & 0.976  & 0.962  & 0.899               & 0.969  & 1.00   & 0.961                \\
\hline
\end{tabular}%
}
\end{table}

\begin{table}[H]
\centering
\caption{XGBoost Results}
\label{tab:xgboost_prior}
\def\arraystretch{1}
\resizebox{\textwidth}{!}{%
\begin{tabular}{|c|ccc|ccc|ccc|ccc|ccc|ccc|} 
\cline{2-19}
\multicolumn{1}{c|}{\textbf{}} & \multicolumn{3}{c|}{\textbf{1 Prior}} & \multicolumn{3}{c|}{\textbf{2 Prior}} & \multicolumn{3}{c|}{\textbf{3 Prior}} & \multicolumn{3}{c|}{\textbf{4 Prior}} & \multicolumn{3}{c|}{\textbf{5 Prior}} & \multicolumn{3}{c|}{\textbf{6 Prior}}  \\ 
\hline
\textbf{Metrics}               & 30 Min & 1 Hour & 4 Hour              & 30min & 1 Hour & 4 Hour               & 30 Min & 1 Hour & 4 Hour              & 30 min & 1 Hour & 4 Hour              & 30min & 1 Hour & 4 Hour               & 30 min & 1 Hour & 4 Hour               \\ 
\hline
Mape                           & 5.97   & 9.01   & 14.83               & 6.26  & 7.89   & 25.15                & 5.72   & 8.32   & 14.44               & 5.97   & 7.27   & 13.81               & 5.76  & 7.01   & 10.83                & 6.05   & 6.91   & 14.29                \\
MAE                            & 0.036  & 0.044  & 0.068               & 0.038 & 0.041  & 0.063                & 0.036  & 0.039  & 0.05                & 0.035  & 0.037  & 0.066               & 0.035 & 0.036  & 0.057                & 0.036  & 0.038  & 0.061                \\
RMSE                           & 0.115  & 0.123  & 0.142               & 0.121 & 0.121  & 0.141                & 0.123  & 0.113  & 0.105               & 0.116  & 0.109  & 0.149               & 0.119 & 0.106  & 0.120                & 0.117  & 0.113  & 0.130                \\
R-Square                       & 0.985  & 0.984  & 0.978               & 0.984 & 0.984  & 0.976                & 0.984  & 0.987  & 0.987               & 0.984  & 0.987  & 0.975               & 0.985 & 0.987  & 0.982                & 0.985  & 0.987  & 0.981                \\
STD of Predicted              & 0.971  & 0.981  & 0.957               & 0.976 & 0.980  & 0.929                & 0.975  & 0.993  & 0.934               & 0.975  & 0.956  & 0.948               & 0.963 & 0.956  & 0.891                & 0.961  & 1.00   & 0.948                \\
STD of Test Set                & 0.973  & 0.983  & 0.971               & 0.982 & 0.980  & 0.929                & 0.985  & 1.00   & 0.948               & 0.985  & 0.962  & 0.956               & 0.976 & 0.962  & 0.899                & 0.969  & 1.00   & 0.961                \\
\hline
\end{tabular}%
}
\end{table}

\begin{table}[H]
\centering
\caption{MLP Results}
\label{tab:mlp_prior}
\def\arraystretch{1}
\resizebox{\textwidth}{!}{%
\begin{tabular}{|c|ccc|ccc|ccc|ccc|ccc|ccc|} 
\cline{2-19}
\multicolumn{1}{c|}{\textbf{}} & \multicolumn{3}{c|}{\textbf{1 Prior}} & \multicolumn{3}{c|}{\textbf{2 Prior}} & \multicolumn{3}{c|}{\textbf{3 Prior}} & \multicolumn{3}{c|}{\textbf{4 Prior}} & \multicolumn{3}{c|}{\textbf{5 Prior}} & \multicolumn{3}{c|}{\textbf{6 Prior}}  \\ 
\hline
\textbf{Metrics}               & 30 Min & 1 Hour & 4 Hour              & 30min & 1 Hour & 4 Hour               & 30 Min & 1 Hour & 4 Hour              & 30 min & 1 Hour & 4 Hour              & 30min & 1 Hour & 4 Hour               & 30 min & 1 Hour & 4 Hour               \\ 
\hline
Mape                           & 9.60   & 12.63  & 22.75               & 8.06  & 10.17  & 21.92                & 10.13  & 10.05  & 19.09               & 8.62   & 10.14  & 18.65               & 8.04  & 11.10  & 14.81                & 9.71   & 8.55   & 18.88                \\
MAE                            & 0.057  & 0.063  & 0.077               & 0.046 & 0.054  & 0.076                & 0.057  & 0.055  & 0.079               & 0.046  & 0.050  & 0.092               & 0.049 & 0.054  & 0.085                & 0.050  & 0.050  & 0.094                \\
RMSE                           & 0.118  & 0.125  & 0.143               & 0.115 & 0.125  & 0.133                & 0.128  & 0.126  & 0.138               & 0.116  & 0.124  & 0.176               & 0.127 & 0.116  & 0.150                & 0.126  & 0.133  & 0.184                \\
R-Square                       & 0.985  & 0.983  & 0.977               & 0.986 & 0.983  & 0.979                & 0.983  & 0.984  & 0.978               & 0.985  & 0.983  & 0.966               & 0.982 & 0.985  & 0.972                & 0.983  & 0.982  & 0.963                \\
STD of  Predicted              & 0.960  & 0.990  & 0.966               & 0.962 & 0.986  & 0.933                & 1.01   & 1.00   & 0.917               & 0.971  & 0.947  & 0.977               & 0.957 & 0.957  & 0.888                & 0.964  & 1.02   & 0.949                \\
STD of Test Set                & 0.973  & 0.983  & 0.971               & 0.982 & 0.980  & 0.929                & 0.985  & 1.00   & 0.948               & 0.981  & 0.962  & 0.956               & 0.976 & 0.962  & 0.899                & 0.969  & 1.00   & 0.961                \\
\hline
\end{tabular}%
}
\end{table}

\begin{table}[H]
\centering
\caption{SVR Results}
\label{tab:svr_prior}
\def\arraystretch{1}
\resizebox{\textwidth}{!}{%
\begin{tabular}{|c|ccc|ccc|ccc|ccc|ccc|ccc|} 
\cline{2-19}
\multicolumn{1}{c|}{\textbf{}} & \multicolumn{3}{c|}{\textbf{1 Prior}} & \multicolumn{3}{c|}{\textbf{2 Prior}} & \multicolumn{3}{c|}{\textbf{3 Prior}} & \multicolumn{3}{c|}{\textbf{4 Prior}} & \multicolumn{3}{c|}{\textbf{5 Prior}} & \multicolumn{3}{c|}{\textbf{6 Prior}}  \\ 
\hline
\textbf{Metrics}               & 30 Min & 1 Hour & 4 Hour              & 30min & 1 Hour & 4 Hour               & 30 Min & 1 Hour & 4 Hour              & 30 min & 1 Hour & 4 Hour              & 30min & 1 Hour & 4 Hour               & 30 min & 1 Hour & 4 Hour               \\ 
\hline
Mape                           & 10.80  & 15.63  & 26.92               & 10.83 & 13.56  & 26.98                & 11.09  & 14.33  & 26.69               & 11.71  & 15.42  & 25.40               & 11.59 & 14.17  & 30.24                & 11.80  & 16.20  & 34.63                \\
MAE                            & 0.080  & 0.086  & 0.110               & 0.083 & 0.090  & 0.112                & 0.085  & 0.092  & 0.114               & 0.086  & 0.092  & 0.148               & 0.090 & 0.095  & 0.155                & 0.092  & 0.111  & 0.177                \\
RMSE                           & 0.139  & 0.143  & 0.174               & 0.140 & 0.153  & 0.169                & 0.147  & 0.157  & 0.180               & 0.149  & 0.160  & 0.237               & 0.160 & 0.168  & 0.257                & 0.163  & 0.209  & 0.322                \\
R-Square                       & 0.979  & 0.978  & 0.967               & 0.979 & 0.975  & 0.966                & 0.977  & 0.975  & 0.963               & 0.976  & 0.972  & 0.938               & 0.972 & 0.969  & 0.917                & 0.971  & 0.956  & 0.887                \\
STD of  Predicted              & 0.961  & 0.970  & 0.932               & 0.963 & 0.950  & 0.890                & 0.957  & 0.964  & 0.902               & 0.946  & 0.930  & 0.897               & 0.938 & 0.918  & 0.810                & 0.926  & 0.932  & 0.817                \\
STD of Test Set                & 0.973  & 0.983  & 0.971               & 0.982 & 0.980  & 0.929                & 0.985  & 1.00   & 0.948               & 0.981  & 0.962  & 0.956               & 0.976 & 0.962  & 0.899                & 0.969  & 1.00   & 0.961                \\
\hline
\end{tabular}%
}
\end{table}

\begin{table}[H]
\centering
\caption{RF Results}
\label{tab:rf_prior}
\def\arraystretch{1}
\resizebox{\textwidth}{!}{%
\begin{tabular}{|c|ccc|ccc|ccc|ccc|ccc|ccc|} 
\cline{2-19}
\multicolumn{1}{c|}{\textbf{}} & \multicolumn{3}{c|}{\textbf{1 Prior}} & \multicolumn{3}{c|}{\textbf{2 Prior}} & \multicolumn{3}{c|}{\textbf{3 Prior}} & \multicolumn{3}{c|}{\textbf{4 Prior}} & \multicolumn{3}{c|}{\textbf{5 Prior}} & \multicolumn{3}{c|}{\textbf{6 Prior}}  \\ 
\hline
\textbf{Metrics}               & 30 Min & 1 Hour & 4 Hour              & 30min & 1 Hour & 4 Hour               & 30 Min & 1 Hour & 4 Hour              & 30 min & 1 Hour & 4 Hour              & 30min & 1 Hour & 4 Hour               & 30 min & 1 Hour & 4 Hour               \\ 
\hline
Mape                           & 5.27   & 8.52   & 16.27               & 5.44  & 7.50   & 20.33                & 5.46   & 7.33   & 14.55               & 5.86   & 7.49   & 13.99               & 5.27  & 6.59   & 11.93                & 5.77   & 7.18   & 14.79                \\
MAE                            & 0.032  & 0.039  & 0.068               & 0.033 & 0.038  & 0.062                & 0.033  & 0.037  & 0.056               & 0.033  & 0.036  & 0.074               & 0.032 & 0.034  & 0.063                & 0.035  & 0.038  & 0.071                \\
RMSE                           & 0.110  & 0.113  & 0.138               & 0.111 & 0.111  & 0.123                & 0.113  & 0.104  & 0.110               & 0.111  & 0.106  & 0.152               & 0.108 & 0.098  & 0.122                & 0.112  & 0.114  & 0.142                \\
R-Square                       & 0.987  & 0.986  & 0.979               & 0.987 & 0.986  & 0.982                & 0.986  & 0.988  & 0.986               & 0.987  & 0.987  & 0.974               & 0.987 & 0.986  & 0.981                & 0.986  & 0.986  & 0.978                \\
STD of  Predicted              & 0.967  & 0.977  & 0.942               & 0.971 & 0.972  & 0.913                & 0.974  & 0.991  & 0.926               & 0.971  & 0.953  & 0.937               & 0.963 & 0.952  & 0.871                & 0.958  & 0.997  & 0.913                \\
STD of Test Set                & 0.973  & 0.983  & 0.971               & 0.982 & 0.980  & 0.929                & 0.985  & 1.00   & 0.948               & 0.981  & 0.962  & 0.956               & 0.976 & 0.962  & 0.899                & 0.969  & 1.00   & 0.961                \\
\hline
\end{tabular}%
}
\end{table}

\end{document}